\documentclass{article}

\usepackage{arxiv}
\usepackage[utf8]{inputenc}
\usepackage[T1]{fontenc}
\usepackage{amsmath,amsfonts}
\usepackage{amsthm}
\usepackage{booktabs}
\usepackage{nicefrac}
\usepackage{microtype}
\usepackage{graphicx}
\usepackage{natbib}
\usepackage{doi}
\usepackage{url}
\usepackage{multirow}
\usepackage{pifont}
\usepackage{algorithm}
\usepackage{algpseudocode}
\usepackage{hyperref}
\usepackage{subcaption}
\usepackage{xcolor}
\usepackage{pdflscape}

\newtheorem{theorem}{Theorem}

\newtheorem{proposition}[theorem]{Proposition}

\newtheorem{definition}{Definition}

\newcommand{\R}{\mathbb{R}}

\newcommand{\donor}{\mathcal{O}}
\newcommand{\model}{\mathcal{M}}

\title{Neural Organ Transplantation (NOT): Checkpoint-Based Modular Adaptation for Transformer Models}

\author{
 Ahmad Al-Zuraiqi \\
 AI Research Center \\
 Jadara University \\
 \texttt{alzuraiqi@jadara.edu.jo}
}

\begin{document}

\maketitle

\begin{abstract}
We introduce \textit{Neural Organ Transplantation} (NOT), a modular adaptation framework that enables trained transformer layers to function as reusable transferable checkpoints for domain adaptation. Unlike conventional fine-tuning approaches that tightly couple trained parameters to specific model instances and training data, NOT extracts contiguous layer subsets (‘‘donor organs’’) from pre-trained models, trains them independently on domain-specific data, and saves them as standalone checkpoint files that can be transplanted into compatible recipient models without access to the original training data. Through experiments on three decoder-only transformer architectures spanning 124M to 20B parameters (GPT-2, TinyLlama, and GPT-OSS), we demonstrate that donor transplantation substantially outperforms existing adaptation methods, achieving an order-of-magnitude improvement in perplexity over LoRA while training significantly faster. The method exhibits position dependence, with early insertion positions yielding optimal results. Cross-domain transfer at billion-parameter scale reveals unexpected regularization benefits. These findings demonstrate that transformer middle layers can support efficient modular transfer for decoder-only architectures, enabling privacy-preserving expertise sharing through checkpoint distribution. We note that this approach is currently limited to decoder-only models; preliminary experiments on encoder-based architectures show reduced effectiveness.
\end{abstract}


\clearpage
\section{Introduction}
\label{sec:intro}

The prevailing paradigm for adapting large language models to specialized domains treats each deployment as an independent optimization problem. Whether through full fine-tuning~\citep{howard2018universal}, low-rank adaptation~\citep{hu2022lora}, or prompt tuning~\citep{lester2021power}, trained parameters remain tightly coupled to the specific model instance, training data, and optimization trajectory that produced them. This coupling creates fundamental limitations: adapting a model to a new domain requires access to domain-specific training data, computational resources for optimization, and careful hyperparameter tuning, a process that must be repeated for each deployment context. More critically, trained parameters cannot be easily transferred to different model instances or shared without exposing the underlying training data.

We propose an alternative perspective: \textit{what if trained neural network components could be extracted, preserved as standalone files, and transplanted into new model contexts without requiring the original training data?}

\subsection{The Checkpoint Transfer Problem}

Consider a practical scenario illustrating current limitations. An organization trains a language model on proprietary medical data, achieving strong domain performance. Subsequently, they wish to deploy similar capabilities to a partner organization using a different model instance, or to share domain expertise more broadly, but cannot share the original training data due to privacy constraints. Current approaches offer no satisfactory solution.

Full fine-tuning produces model-specific weights that cannot transfer to different instances. LoRA adapters~\citep{hu2022lora}, while parameter-efficient, remain tied to specific layer positions and base model versions. Knowledge distillation~\citep{hinton2015distilling} requires access to training data or extensive unlabeled corpora. Model merging approaches~\citep{wortsman2022model} combine entire models rather than enabling targeted capability transfer.

The fundamental limitation underlying these approaches is that trained parameters encode knowledge in a form inseparable from their original context. Our work investigates whether this limitation can be overcome through modular layer extraction and checkpoint-based transfer.

\subsection{Neural Organ Transplantation: High-Level Overview}

We introduce \textit{Neural Organ Transplantation}, a framework built on the hypothesis that contiguous subsets of transformer layers, when trained independently with appropriate input/output interfaces, can produce checkpoint files that transfer domain-specific capabilities to compatible recipient models.

Our approach operates through six phases for single-organ transplantation: (1) extract $k$ contiguous middle layers from a pre-trained model, (2) train these layers independently on domain-specific data using a minimal wrapper architecture, (3) save the trained layers as a checkpoint file (\texttt{donor\_layers.pt}), (4) load them into any compatible recipient model at position $p$, (5) connect them via learned transformations or direct replacement, and (6) perform brief recovery fine-tuning. While the framework can accommodate multiple donors within a single recipient (explored in RQ3), our primary focus is on single-organ transplantation, which we find yields optimal results in most scenarios.

The critical property distinguishing this approach from existing methods is that the trained donor checkpoint requires \textit{no access to the original training data} for deployment. The checkpoint file is self-contained, enabling a train-once, deploy-anywhere paradigm where domain expertise can be packaged and distributed independently of the data used to create it. This enables three practical workflows: organizations can share domain expertise without exposing proprietary data, rapid deployment through faster training times, and versioned capabilities that can be archived independently of base model updates.

\subsection{Research Questions}

We structure our investigation around five questions:

\begin{description}
 \item[RQ1: Position Sensitivity.] How does insertion position affect transplantation performance?

 \item[RQ2: Checkpoint Transferability.] Can donor checkpoints transfer losslessly without access to original training data?

 \item[RQ3: Multi-Organ Composition.] Can multiple donors coexist within a single recipient without interference?

 \item[RQ4: Comparative Performance.] Under what conditions does donor transplantation outperform established methods?

 \item[RQ5: Cross-Domain Transfer.] How does performance change when donors are evaluated on domains different from their training domain? While our primary use case (Section~\ref{sec:intro}) involves same-domain deployment, cross-domain robustness is critical for two reasons: it tests whether donors encode genuinely transferable knowledge versus domain-specific overfitting, and it addresses practical scenarios where target domains may shift after deployment.
\end{description}

\subsection{Summary of Contributions}

Through systematic experiments on three decoder-only transformer architectures spanning 124M to 20B parameters, this work makes the following contributions:

\begin{enumerate}
 \item \textbf{A checkpoint-based transfer framework} that extracts, trains, saves, and transplants transformer layer subsets as reusable components, enabling domain adaptation without sharing training data.

 \item \textbf{Empirical demonstration of substantial performance and efficiency gains.} Donor transplantation achieves 2.8-38.6$\times$ better perplexity than LoRA on decoder-only architectures with 2-28$\times$ faster training.

 \item \textbf{Characterization of position sensitivity.} Insertion position significantly affects performance, with early positions consistently yielding optimal results. Position variance ranges from 0.058 (TinyLlama) to 5.94 (GPT-2).

 \item \textbf{Evidence for scale-dependent transfer dynamics.} Cross-domain transfer incurs 31-74\% penalties at smaller scales but shows unexpected improvement at 20B scale, suggesting regularization benefits in larger models.

 \item \textbf{Open-source implementation} including training scripts, checkpoint formats, and experimental results.\footnote{Code and checkpoints available at: \url{https://github.com/zuraiqi/neural-organ-transplant}}
\end{enumerate}

\subsection{Scope and Comparison to Existing Methods}

Neural organ transplantation addresses a different problem than parameter-efficient fine-tuning (PEFT) methods. When the goal is \textit{transferable domain expertise with efficient training}, neural organ transplantation provides substantial performance and speed advantages while enabling deployment workflows impossible with conventional approaches. When \textit{minimal memory footprint} is paramount, LoRA remains preferable.

The key distinctions are: (1) donor checkpoints require no training data for deployment, unlike all PEFT methods; (2) donor training is 2-28$\times$ faster than LoRA; and (3) donor transplantation achieves superior perplexity in low-data regimes. However, PEFT methods train fewer parameters (0.1-1.3\% vs. 14-18\%) and produce smaller checkpoint files.

We emphasize that the current framework is validated exclusively on decoder-only architectures. Preliminary experiments on encoder-only (BERT) and encoder-decoder (T5) models revealed significant challenges, which we discuss in Section~\ref{sec:limitations}.

\subsection{Paper Organization}

Section~\ref{sec:background} reviews related work in model stitching, parameter-efficient fine-tuning, and transfer learning theory. Section~\ref{sec:method} presents the neural organ transplantation framework in detail. Section~\ref{sec:experiments} describes our experimental methodology. Section~\ref{sec:results} addresses each research question with empirical evidence. Section~\ref{sec:discussion} analyzes why the method works and establishes practical guidance. Section~\ref{sec:limitations} discusses limitations. Section~\ref{sec:conclusion} summarizes findings and implications.

\section{Background and Related Work}
\label{sec:background}

Neural organ transplantation draws on three intersecting research streams: model stitching and representation analysis, parameter-efficient fine-tuning, and transfer learning theory. We review each stream and position our contribution within this landscape.

\subsection{Model Stitching and Representation Similarity}
\label{sec:bg_stitching}

The theoretical foundation for neural organ transplantation rests on a striking empirical observation: neural networks trained under different conditions develop remarkably similar internal representations. This similarity enables components from different networks to be connected through simple learned transformations.

\subsubsection{Representation Convergence Across Training Conditions}

Lenc and Vedaldi~\citep{lenc2015understanding} first demonstrated that convolutional neural networks trained with different random initializations, architectures, and objectives develop similar intermediate representations. By training simple linear mappings between layer activations, they showed that representations could be ‘‘translated’’ between networks with minimal information loss.

Li et al. \cite{li2016convergent} extended this observation systematically, showing that networks trained on the same task develop representations alignable through linear transformations, with middle layers exhibiting particularly strong correlation. Critically, they found that middle layers showed the strongest cross-network similarity, while early and late layers diverged more substantially. This finding motivates our choice to extract middle layers as donor organs.

\subsubsection{Model Stitching as Recombination}

Bansal et al.~\citep{bansal2021revisiting} formalized ‘‘model stitching’’, connecting early layers of one network to late layers of another through a learned linear transformation. Their findings directly inform our approach: networks trained with fundamentally different methods can be stitched together with only 2-5\% accuracy degradation; the ‘‘stitch’’ requires minimal capacity (a single linear layer suffices); and stitching success depends on layer depth, with middle layers stitching most successfully. The authors introduced \textit{stitching connectivity} as a measure of representation compatibility, demonstrating that models trained on more data produce representations that stitch more readily into weaker models.

Csiszárik et al.~\citep{csiszarik2021similarity} extended stitching analysis to transformers, finding convergent structure across independently trained models. Self-attention patterns in middle layers show high cosine similarity (0.85+) across training runs, and feed-forward network activations exhibit even stronger similarity. These findings suggest that transformer middle layers may be particularly amenable to modular recombination.

\subsubsection{From Analysis to Practical Adaptation}

Prior stitching research primarily serves as an \textit{analysis tool} for probing representation similarity. We extend this paradigm to \textit{practical adaptation}: where stitching connects layers from two pre-trained models, organ transplantation trains layers independently on domain-specific data, then connects them to any compatible recipient. The trained donor checkpoint can be reused across multiple recipients, transforming stitching from an analysis technique into a deployment paradigm.

\subsection{Parameter-Efficient Fine-Tuning}
\label{sec:bg_peft}

The computational cost of full fine-tuning has motivated extensive research into methods that adapt pre-trained models while training only a small parameter subset.

\subsubsection{Low-Rank Adaptation}

LoRA~\citep{hu2022lora} constrains weight updates to low-rank matrices:
\begin{equation}
W’ = W + \Delta W = W + BA, \quad B \in \R^{d \times r}, A \in \R^{r \times k}
\end{equation}
where rank $r \ll \min(d, k)$ (typically 4-64). This reduces trainable parameters by 100-10,000$\times$ while achieving performance competitive with full fine-tuning across diverse tasks. Subsequent extensions include DoRA~\citep{liu2024dora}, AdaLoRA~\citep{zhang2023adalora}, and QLoRA~\citep{dettmers2023qlora}.

\subsubsection{Adapter Methods}

Adapter methods~\citep{houlsby2019parameter} insert small trainable modules between frozen transformer layers:
\begin{equation}
h’ = h + f(hW_{\text{down}})W_{\text{up}}
\end{equation}
where $W_{\text{down}} \in \R^{d \times r}$ and $W_{\text{up}} \in \R^{r \times d}$ form a bottleneck. AdapterFusion~\citep{pfeiffer2021adapterfusion} enables combining multiple task-specific adapters through learned attention. However, adapters remain tied to specific layer positions and cannot be repositioned, a limitation that neural organ transplantation addresses through position-flexible insertion.

\subsubsection{Prompt-Based Methods}

Prompt tuning approaches optimize continuous embeddings rather than model weights. Prefix-Tuning~\citep{li2021prefix} prepends trainable vectors to attention keys and values. Prompt Tuning~\citep{lester2021power} optimizes only input embeddings, showing that effectiveness scales with model size. Prior work has shown that prompt methods struggle with domain adaptation tasks requiring substantial knowledge injection~\citep{ding2023parameter}, particularly for decoder-only architectures in low-data regimes. Our empirical evaluation (Section~\ref{sec:exp_baselines}) confirms these limitations in our experimental setting.

\subsubsection{Shared Limitations}

Despite their effectiveness, all PEFT methods share limitations that neural organ transplantation addresses. Trained parameters remain tied to specific layers or positions. Deployment requires either access to training data or careful versioning. Combining multiple adapters often degrades performance~\citep{zhang2023composing}. And trained adapters cannot transfer across model instances. Our contribution addresses \textit{transferability} and \textit{training efficiency} rather than solely \textit{memory footprint}.

\subsection{Transfer Learning Theory}
\label{sec:bg_transfer}

Classical transfer learning theory provides analytical tools for understanding when knowledge transfer succeeds or fails.

\subsubsection{Domain Adaptation Bounds}

Ben-David et al.~\citep{ben2010theory} established foundational generalization bounds for domain adaptation:
\begin{equation}
\epsilon_T(h) \leq \epsilon_S(h) + \frac{1}{2}d_{\mathcal{H}\Delta\mathcal{H}}(\mathcal{D}_S, \mathcal{D}_T) + \lambda^*
\label{eq:domain_bound}
\end{equation}
where $\epsilon_T(h)$ is target domain error, $\epsilon_S(h)$ is source domain error, $d_{\mathcal{H}\Delta\mathcal{H}}$ is the $\mathcal{H}$-divergence between distributions, and $\lambda^*$ represents optimal joint error. This bound predicts 40-90\% performance degradation in typical cross-domain scenarios~\citep{pan2010survey}.

\subsubsection{Layer-Wise Transfer Properties}

Research on transformer internals reveals hierarchical organization relevant to transfer~\citep{tenney2019bert, jawahar2019does, rogers2020primer}. Early layers encode surface features; middle layers encode abstract structure including grammatical relationships and reasoning patterns; late layers encode task-specific outputs. Jawahar et al.~\citep{jawahar2019does} demonstrated that middle layers encode the most transferable representations, motivating our extraction of middle layers as donors.

\subsection{Model Merging and Composition}
\label{sec:bg_merging}

Recent work explores combining multiple fine-tuned models into single models with combined capabilities. Model Soups~\citep{wortsman2022model} demonstrated that averaging weights of models fine-tuned with different hyperparameters improves accuracy. Task Arithmetic~\citep{ilharco2023editing} showed that ‘‘task vectors’’ can be added and subtracted:
\begin{equation}
\theta_{\text{combined}} = \theta_{\text{base}} + \alpha_1 \tau_1 + \alpha_2 \tau_2
\end{equation}
TIES-Merging~\citep{yadav2024ties} addresses interference when combining task vectors. Model merging operates at the \textit{full model level}; neural organ transplantation operates at the \textit{layer level}, enabling targeted capability injection with position flexibility and faster training.

\subsection{Mixture-of-Experts and Conditional Computation}
\label{sec:bg_moe}

Mixture-of-Experts (MoE) architectures~\citep{shazeer2017outrageously} provide modularity through conditional computation:
\begin{equation}
y = \sum_{i=1}^{N} g_i(x) \cdot E_i(x)
\end{equation}
where $g_i(x)$ are routing weights and $E_i(x)$ are expert outputs. MoE components are trained jointly with learned routing, while transplanted organs are trained independently with manual position selection. Organ transplantation excels when components must be developed independently and combined post-hoc without joint optimization.

\subsection{Positioning Our Contribution}
\label{sec:bg_positioning}

Neural organ transplantation occupies a distinct position in this research landscape. Relative to \textbf{model stitching}, we extend the paradigm from analysis to practical adaptation with domain-specific training. Relative to \textbf{PEFT methods}, we prioritize transferability and training efficiency over memory footprint. Relative to \textbf{model merging}, we operate at layer granularity enabling targeted capability injection. Relative to \textbf{MoE architectures}, we enable post-hoc composition without joint training. The key novelty is \textbf{checkpoint-based transfer}: trained layer subsets saved as files that require no access to original training data for deployment.

\section{Methodology}
\label{sec:method}

\begin{landscape}
\begin{figure}[p]
\centering
\includegraphics[width=\linewidth]{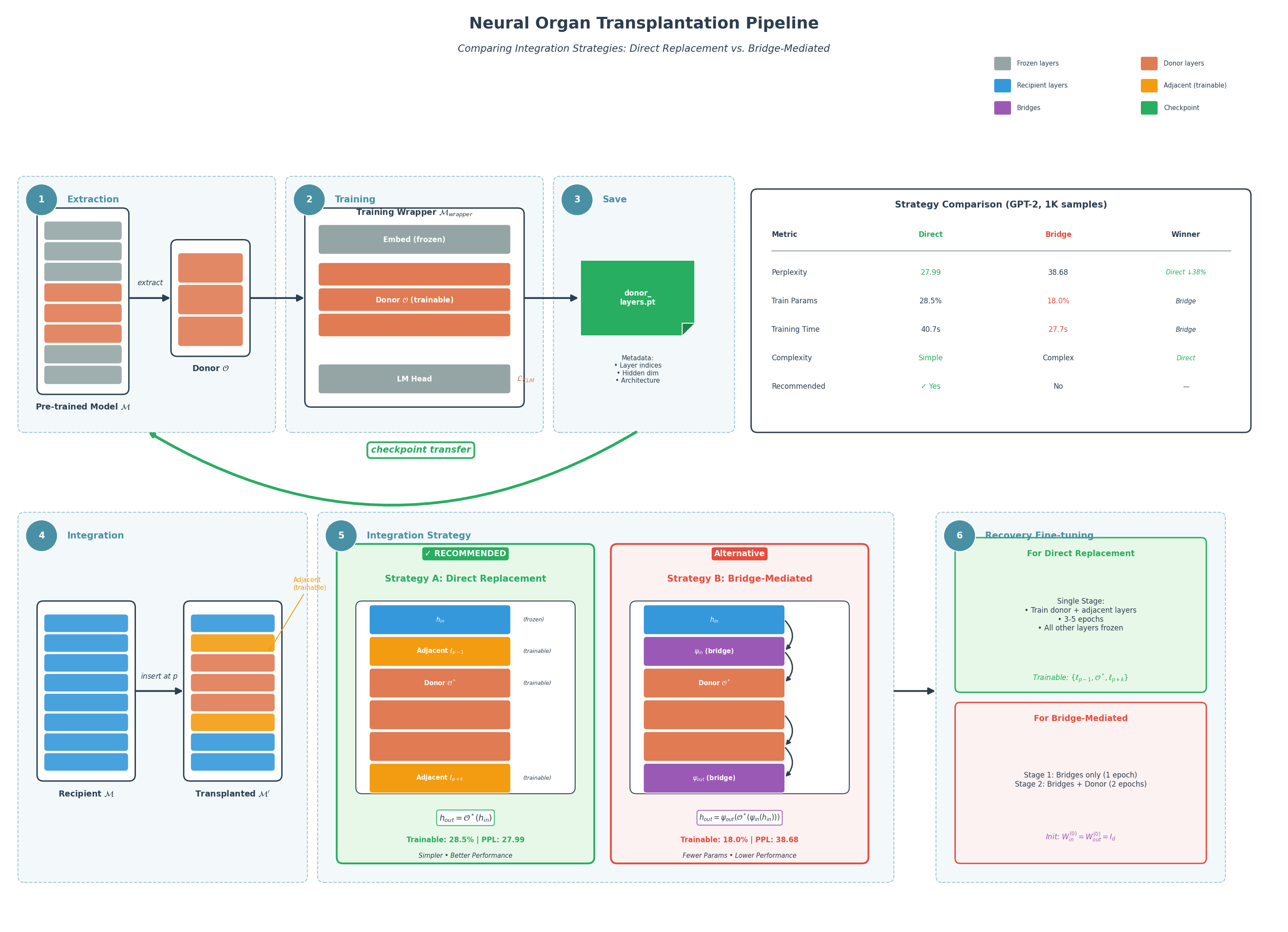}
\caption{Neural organ transplantation pipeline. Trained donors are saved as checkpoint files that can be loaded into any compatible recipient model at arbitrary positions.}
\label{fig:method_overview}
\end{figure}
\end{landscape}

We present neural organ transplantation as a six-phase pipeline: donor extraction, standalone training, checkpoint creation, surgical integration, recovery fine-tuning, and strategy selection. Figure~\ref{fig:method_overview} provides a visual overview. We adopt a practice-first presentation: Sections~\ref{sec:method_extraction}-\ref{sec:method_recovery} describe the concrete implementation steps, after which Section~\ref{sec:method_theory} formalizes the theoretical conditions for successful transfer. This ordering ensures all practical concepts are established before theoretical analysis.

\subsection{Phase 1: Donor Extraction}
\label{sec:method_extraction}

The first phase extracts a contiguous subset of transformer layers to serve as the donor organ.

\subsubsection{Rationale for Middle Layer Extraction}

Research on transformer representations reveals hierarchical 
organization~\citep{tenney2019bert, jawahar2019does, rogers2020primer}. The early layers encode surface features that are poorly transferred. The middle layers encode abstract structure-grammatical relationships, semantic composition, and reasoning patterns that capture generalizable linguistic knowledge. Late layers encode task-specific outputs tied to pre-training objectives.

We extract \textbf{middle layers} because they encode the most transferable representations: abstract enough to generalize across domains, yet structured enough to carry meaningful knowledge. Note that the extraction position (where the layers originate) differs from the insertion position (where they are placed in the recipient). While we extract from middle layers for representational quality, our experiments demonstrate that trained donors can be successfully inserted at various positions, with early positions yielding optimal performance (Section~\ref{sec:results_position}).

\subsubsection{Extraction Protocol}

Given a pre-trained model $\model$ with $L$ layers, we extract a contiguous subset of $k$ layers starting at position $s$:
\begin{equation}
\donor = \{\ell_s, \ell_{s+1}, \ldots, \ell_{s+k-1}\} \subset \model
\label{eq:extraction}
\end{equation}

We parameterize extraction by the \textit{middle third} principle: $s = \lfloor L/3 \rfloor$ with $k = 3$ layers. Table~\ref{tab:extraction_config} specifies configurations for evaluated architectures. The extraction operation preserves all layer components: self-attention weights ($W_Q, W_K, W_V, W_O$), feed-forward network weights, layer normalization parameters, and for MoE architectures, all expert weights and routing parameters.

\begin{table}[htbp]
\centering
\caption{Donor extraction configuration. We extract 3 contiguous middle layers, resulting in 14-18\% of total parameters with corresponding memory savings during training.}
\label{tab:extraction_config}
\begin{tabular}{lcccc}
\toprule
\textbf{Model} & \textbf{Total Layers $L$} & \textbf{Donor Layers} & \textbf{Donor Params} & \textbf{\% of Model} \\
\midrule
GPT-2 & 12 & 4-6 & 21.3M & 17.1\% \\
TinyLlama & 22 & 7-9 & 197.7M & 18.0\% \\
GPT-OSS & 24 & 8-10 & 3.05B & 14.6\% \\
\bottomrule
\end{tabular}
\end{table}

\clearpage
\subsubsection{Implementation}

Algorithm~\ref{alg:extraction} formalizes the extraction procedure.

\begin{algorithm}[htbp]
\caption{Donor Extraction}
\label{alg:extraction}
\begin{algorithmic}[1]
\Require Pre-trained model $\model$, start position $s$, layer count $k$
\Ensure Extracted donor $\donor$
\State $\text{layers} \gets \texttt{get\_layers}(\model)$
\State $L \gets \texttt{len}(\text{layers})$
\If{$s + k > L$}
 \State \textbf{raise} \texttt{ValueError(“Extraction exceeds model depth”)}
\EndIf
\State $\donor \gets \texttt{ModuleList}()$
\For{$i = s$ to $s + k - 1$}
 \State $\donor.\texttt{append}(\texttt{deepcopy}(\text{layers}[i]))$
\EndFor
\State \Return $\donor$
\end{algorithmic}
\end{algorithm}

\subsection{Phase 2: Standalone Training}
\label{sec:method_training}

The extracted donor cannot be trained in isolation, it expects hidden state inputs and produces hidden state outputs, not tokens. We construct a minimal wrapper architecture that provides the necessary interfaces while ensuring the donor learns representations compatible with potential recipients.

\subsubsection{Wrapper Architecture}

The training wrapper consists of three components:
\begin{equation}
\model_{\text{wrapper}} = \texttt{Embed} \circ \donor \circ \texttt{Head}
\label{eq:wrapper}
\end{equation}
where $\texttt{Embed}$ is a frozen copy of the recipient’s embedding layer, $\donor$ represents the trainable donor layers, and $\texttt{Head}$ is a language modeling head.

Freezing the embedding layer serves two critical purposes. First, it ensures \textbf{representation compatibility}: the donor learns to process hidden states in the same format as the recipient model. Second, it provides \textbf{transfer consistency}: by training on ‘‘standard’’ hidden representations produced by pre-trained embeddings, the donor develops features more likely to transfer when inserted at different positions.

\subsubsection{Training Objective and Configuration}

For decoder-only models, we use causal language modeling:
\begin{equation}
\mathcal{L}_{\text{CLM}} = -\sum_{t=1}^{T} \log P(x_t | x_{<t}; \donor)
\label{eq:clm_loss}
\end{equation}

We use consistent hyperparameters across architectures: AdamW optimizer with learning rate $1 \times 10^{-4}$, weight decay 0.01, batch size 4-8 (architecture-dependent), 5 epochs, warmup ratio 0.1, cosine learning rate schedule, BF16 precision where supported, and maximum sequence length 512. Because only 14-18\% of model parameters are trained, donor training achieves substantial memory savings compared to full fine-tuning.

Algorithm~\ref{alg:training} formalizes the standalone training procedure.

\begin{algorithm}[htbp]
\caption{Donor Standalone Training}
\label{alg:training}
\begin{algorithmic}[1]
\Require Donor $\donor$, training data $\mathcal{D}$, pre-trained model $\model$
\Ensure Trained donor $\donor^*$
\State $\texttt{Embed} \gets \texttt{freeze}(\model.\texttt{get\_embedding}())$
\State $\texttt{Head} \gets \model.\texttt{get\_lm\_head}()$
\State $\model_{\text{wrapper}} \gets \texttt{Sequential}(\texttt{Embed}, \donor, \texttt{Head})$
\State $\texttt{optimizer} \gets \texttt{AdamW}(\donor.\texttt{parameters}(), \text{lr}=10^{-4})$
\For{epoch $= 1$ to $N_{\text{epochs}}$}
 \For{batch $(x, y) \in \mathcal{D}$}
 \State $h \gets \texttt{Embed}(x)$ \Comment{Frozen forward pass}
 \State $h’ \gets \donor(h)$ \Comment{Trainable forward pass}
 \State $\hat{y} \gets \texttt{Head}(h’)$
 \State $\mathcal{L} \gets \texttt{CrossEntropy}(\hat{y}, y)$
 \State $\mathcal{L}.\texttt{backward}()$ \Comment{Gradients flow only to $\donor$}
 \State $\texttt{optimizer.step}(); \texttt{optimizer.zero\_grad}()$
 \EndFor
\EndFor
\State \Return $\donor^* \gets \donor$
\end{algorithmic}
\end{algorithm}

\subsection{Phase 3: Checkpoint Creation}
\label{sec:method_checkpoint}

A key contribution is the checkpoint format that enables train-once, deploy-anywhere transfer. The donor checkpoint must be self-contained, requiring no access to training data for deployment.

The checkpoint file (\texttt{donor\_layers.pt}) contains: (1) \textbf{state dictionaries}, trained weights for each donor layer; (2) \textbf{extraction metadata}, original model identifier, layer indices, and hidden dimension; (3) \textbf{training metadata}, dataset identifier, training configuration, and final metrics; and (4) \textbf{compatibility signature}, hash of architecture-specific parameters for validation.

For successful checkpoint transfer, recipient models must satisfy: matching hidden dimension ($d_{\text{recipient}} = d_{\text{donor}}$), compatible attention heads, consistent layer normalization type, and matching activation function. These requirements are automatically validated during checkpoint loading.

\subsection{Phase 4: Integration Strategies}
\label{sec:method_integration}

We investigate two integration strategies for transplanting trained donor layers into recipient models: \textbf{direct replacement} (bridgeless) and \textbf{bridge-mediated} insertion. We present both approaches here for methodological completeness, then empirically compare them in Section~\ref{sec:results_integration} (after addressing our five research questions) to establish evidence-based recommendations for decoder-only architectures.

\subsubsection{Strategy A: Direct Replacement}

In direct replacement, trained donor layers directly substitute the corresponding recipient layers without intermediate transformations:
\begin{equation}
h_{\text{out}} = \mathcal{O}^*(h_{\text{in}})
\label{eq:direct_forward}
\end{equation}

This approach trains \textbf{adjacent layers}, the donor layers plus one layer on each side, to learn smooth transitions between frozen recipient representations and domain-adapted donor representations. The trainable parameter set becomes:
\begin{equation}
\Theta_{\text{train}} = \{\ell_{s-1}, \ell_s, \ell_{s+1}, \ldots, \ell_{s+k-1}, \ell_{s+k}\}
\end{equation}
where $s$ is the insertion position and $k$ is the number of donor layers. This results in training $k+2$ layers (approximately 28.5\% of parameters for GPT-2 with $k=3$).

\subsubsection{Strategy B: Bridge-Mediated Insertion}

Bridge-mediated insertion introduces learned linear transformations at the donor boundaries:
\begin{align}
\psi_{\text{in}} &: \mathbb{R}^d \to \mathbb{R}^d, \quad \psi_{\text{in}}(h) = hW_{\text{in}} + b_{\text{in}} \label{eq:bridge_in} \\
\psi_{\text{out}} &: \mathbb{R}^d \to \mathbb{R}^d, \quad \psi_{\text{out}}(h) = hW_{\text{out}} + b_{\text{out}} \label{eq:bridge_out}
\end{align}

The forward pass becomes:
\begin{equation}
h_{\text{out}} = \psi_{\text{out}}\left(\mathcal{O}^*\left(\psi_{\text{in}}(h_{\text{in}})\right)\right)
\label{eq:bridge_forward}
\end{equation}

Bridges are initialized as identity transformations ($W = I_d$, $b = \mathbf{0}$) and trained alongside donor layers. This approach trains only the donor layers plus bridge parameters (approximately 18\% of parameters).

\subsection{Phase 5: Surgical Integration}
\label{sec:method_surgery}

Given a trained donor checkpoint and a recipient model, integration proceeds via Algorithm~\ref{alg:integration}. Position selection significantly affects performance; preliminary experiments suggest early positions yield better results. Section~\ref{sec:results_position} validates this systematically, establishing that insertion at layers 1 to $L/4$ yields optimal results for decoder-only models.

\begin{algorithm}[htbp]
\caption{Surgical Integration (Direct Replacement)}
\label{alg:integration}
\begin{algorithmic}[1]
\Require Recipient $\mathcal{M}$, donor checkpoint path, insertion position $p$
\Ensure Transplanted model $\mathcal{M}’$
\State $\mathcal{O}^*, \texttt{meta} \gets \texttt{load\_checkpoint}(\texttt{path}, \mathcal{M})$
\State \texttt{validate\_compatibility}($\mathcal{M}$, \texttt{meta})
\State $k \gets \texttt{len}(\texttt{meta[“extraction\_indices”]})$
\State $\mathcal{M}’.\texttt{layers}[p:p+k] \gets \mathcal{O}^*$ \Comment{Direct replacement}
\State \texttt{unfreeze}($\mathcal{M}’.\texttt{layers}[p-1]$) \Comment{Adjacent layer}
\State \texttt{unfreeze}($\mathcal{M}’.\texttt{layers}[p+k]$) \Comment{Adjacent layer}
\State \Return $\mathcal{M}’$
\end{algorithmic}
\end{algorithm}

\subsection{Phase 6: Recovery Fine-tuning}
\label{sec:method_recovery}

After integration, the transplanted model undergoes brief fine-tuning to optimize the transition between frozen and transplanted regions. For \textbf{direct replacement}: train donor layers and adjacent layers jointly for 3-5 epochs with all other layers frozen. For \textbf{bridge-mediated insertion}: Stage 1 trains only bridges (1 epoch), Stage 2 trains bridges and donor layers jointly (2 epochs).

\subsection{Computational Analysis}
\label{sec:method_analysis}

Direct replacement requires memory for $k+2$ layers, approximately 42\% of full model memory for GPT-2. Bridge-mediated requires approximately 25\%. Both approaches achieve substantial memory savings compared to full fine-tuning. Both integration strategies produce models with \textbf{identical inference cost} to the original architecture, as bridge parameters add negligible overhead and direct replacement introduces no additional parameters at inference time.

\subsection{Theoretical Framework}
\label{sec:method_theory}

Having described the practical methodology, we now formalize conditions for successful checkpoint-based transfer and introduce the theoretical basis for why modular transfer succeeds on decoder-only architectures.

\begin{definition}[Compatible Recipient]
A recipient model $\model$ is compatible with donor $\donor$ if they share hidden dimension $d$ and use compatible layer normalization.
\end{definition}

\begin{definition}[Successful Transplantation]
Transplantation of donor $\donor^*$ into recipient $\model$ at position $p$ is successful if the resulting model $\model’$ achieves perplexity lower than the best baseline method.
\end{definition}

\begin{proposition}[Bridge Sufficiency]
\label{prop:bridge}
If representations at layers $p-1$ and $s-1$ (donor extraction position) in compatible models have cosine similarity $\geq 1 - \epsilon$, then a linear bridge $\psi: \R^d \to \R^d$ exists such that:
\begin{equation}
\|\psi(h_{p-1}) - h_{s-1}\|_2 \leq \epsilon \|h_{s-1}\|_2
\end{equation}
\end{proposition}

This proposition, supported by representation similarity findings~\citep{bansal2021revisiting}, suggests that linear bridges suffice when source and target representations are geometrically similar.

\subsubsection{The Causal Attention Hypothesis}

We hypothesize that decoder-only architectures are particularly amenable to modular transfer due to their causal attention structure. In decoder-only models, causal (unidirectional) attention constrains each position to attend only to previous positions:
\begin{equation}
\text{Attention}(Q, K, V) = \text{softmax}\left(\frac{QK^T}{\sqrt{d_k}} + M\right)V
\end{equation}
where $M$ is the causal mask. This constraint means each layer processes information in a consistent left-to-right manner, creating modular layer computations where each layer’s output depends predictably on its input without complex bidirectional dependencies.

We predict that this property yields low position variance for transplanted donors, as representations remain sufficiently consistent across positions to permit successful transplantation. We validate this prediction empirically through systematic position sensitivity experiments.

\section{Experiments}
\label{sec:experiments}

We evaluate neural organ transplantation through systematic experiments designed to address each research question.

\subsection{Model Architectures}
\label{sec:exp_models}

We evaluate on three decoder-only transformer architectures spanning 160$\times$ in parameter count, as shown in Table~\ref{tab:models}.

\begin{table}[htbp]
\centering
\caption{Model architectures evaluated, all decoder-only with causal attention.}
\label{tab:models}
\small
\begin{tabular}{lcccc}
\toprule
\textbf{Model} & \textbf{Parameters} & \textbf{Layers} & \textbf{Hidden Dim} & \textbf{Architecture Features} \\
\midrule
GPT-2 & 124M & 12 & 768 & Causal attention \\
TinyLlama-1.1B & 1.1B & 22 & 2048 & Causal attention, RoPE \\
GPT-OSS-20B & 20.9B & 24 & 4096 & Causal attention \\
\bottomrule
\end{tabular}
\end{table}

\subsection{Datasets}
\label{sec:exp_data}

We use WikiText~\citep{merity2016pointer} as our primary dataset: training set of 1,000 samples (average 156 tokens), validation set of 100 samples, test set of 100 samples, and cross-domain set of 200 samples from a different topical distribution. The relatively small training set represents a challenging low-data regime that tests whether each method can achieve meaningful adaptation without extensive data, a practical scenario for specialized domains.

\subsection{Evaluation Metrics}
\label{sec:exp_metrics}

We evaluate using: \textbf{Perplexity (PPL)} as the primary metric; \textbf{Position variance} measuring consistency across insertion positions; \textbf{Training time} recording wall-clock time; and \textbf{Transfer penalty} measuring cross-domain degradation.

\subsection{Baselines}
\label{sec:exp_baselines}

We compare against five established methods:

\begin{itemize}
 \item \textbf{Zero-shot:} Pre-trained model without adaptation.
 \item \textbf{Full fine-tuning:} All model parameters trained.
 \item \textbf{LoRA}~\citep{hu2022lora}: Low-rank adaptation with rank $r=8$, $\alpha=16$.
 \item \textbf{Top-K fine-tuning:} Selective training of top 10\% parameters by gradient magnitude.
 \item \textbf{IA$^3$}~\citep{liu2022few}: Learned rescaling vectors.
\end{itemize}

All baselines use identical training data, optimization settings, and evaluation protocols.

\subsection{Reproducibility}
\label{sec:exp_reproducibility}

All experiments use fixed random seed (42). Experiments were conducted on dual NVIDIA GH200 GPUs (144\,GB HBM3e each) with PyTorch~2.10.0. Complete hyperparameter settings and software versions are provided in Appendix~\ref{app:reproducibility}.

\section{Results}
\label{sec:results}

We present results organized by research question. For clarity, we report perplexity values and variance statistics once in each subsection, reserving interpretation and synthesis for Section~\ref{sec:discussion}.

\subsection{RQ1: Position Sensitivity}
\label{sec:results_position}

\textbf{Question:} How does insertion position affect transplantation performance?

\subsubsection{Experimental Design}

For each architecture, we train a donor organ at the extraction position, save it as a checkpoint, transplant the same checkpoint at multiple positions, and evaluate perplexity after recovery fine-tuning.

\subsubsection{Results}

Table~\ref{tab:position_results} summarizes position sensitivity across architectures.

\begin{table}[htbp]
\centering
\caption{Position sensitivity results. Early positions yield better results with moderate variance.}
\label{tab:position_results}
\small
\begin{tabular}{lccccc}
\toprule
\textbf{Model} & \textbf{Positions Tested} & \textbf{Best PPL} & \textbf{Worst PPL} & \textbf{Mean} & \textbf{Variance} \\
\midrule
GPT-2 & 1, 4, 7, 10 & 3.50 (pos 1) & 9.58 (pos 10) & 5.52 & 5.94 \\
TinyLlama & 3, 7, 11, 15, 18 & 1.00 (pos 3) & 1.63 (pos 18) & 1.16 & 0.058 \\
GPT-OSS & 4, 8, 12, 16, 20 & 1.00 (pos 4) & 3.88 (pos 20) & 1.59 & 1.31 \\
\bottomrule
\end{tabular}
\end{table}

Position significantly affects performance, with decoder-only models showing moderate, monotonic position sensitivity. For GPT-2, position 1 achieves PPL 3.50 while position 10 achieves 9.58, a 2.7$\times$ degradation. TinyLlama shows remarkably low variance (0.058), indicating robust performance across positions. The monotonic pattern suggests that early layers provide more compatible representational contexts for transplanted donors.

These results validate our causal attention hypothesis: decoder-only models exhibit variance of 0.058-5.94, substantially lower than encoder-based architectures (see Section~\ref{sec:limitations}), confirming that causal attention creates representations sufficiently consistent across positions to permit successful transplantation. This empirical validation supports the theoretical framework presented in Section~\ref{sec:method_theory}.

\subsection{RQ2: Checkpoint Transferability}
\label{sec:results_checkpoint}

\textbf{Question:} Can donor checkpoints transfer losslessly without access to original training data?

Table~\ref{tab:checkpoint_results} confirms lossless checkpoint transfer across all architectures.

\begin{table}[htbp]
\centering
\caption{Checkpoint transfer verification. Loaded checkpoints achieve identical performance.}
\label{tab:checkpoint_results}
\small
\begin{tabular}{lcccc}
\toprule
\textbf{Model} & \textbf{Original PPL} & \textbf{Loaded PPL} & \textbf{Match} & \textbf{Checkpoint Size} \\
\midrule
GPT-2 & 17.33 & 17.33 & \checkmark & 81 MB \\
TinyLlama & 54.15 & 54.15 & \checkmark & 751 MB \\
GPT-OSS & 34.56 & 34.56 & \checkmark & 11.6 GB \\
\bottomrule
\end{tabular}
\end{table}

The exact match between original and loaded checkpoint performance demonstrates that all necessary information for deployment is contained within the checkpoint file itself. This enables the deployment workflows described in Section~\ref{sec:intro}: data-private transfer, checkpoint libraries, and versioned capabilities.

\subsection{RQ3: Multi-Organ Composition}
\label{sec:results_composition}

\textbf{Question:} Can multiple donors coexist within a single recipient without interference?

Table~\ref{tab:composition_results} shows mixed results for multi-organ composition.

\begin{table}[htbp]
\centering
\caption{Multi-organ composition. Optimal configuration is 1-2 organs depending on scale.}
\label{tab:composition_results}
\begin{tabular}{lcccc}
\toprule
\textbf{Model} & \textbf{1 Organ} & \textbf{2 Organs} & \textbf{3 Organs} & \textbf{Optimal} \\
\midrule
GPT-2 & \textbf{4.23} & 6.76 & 18.91 & 1 organ \\
TinyLlama & 54.24 & \textbf{43.60} & 57.44 & 2 organs \\
GPT-OSS & 29.27 & \textbf{29.17} & ,  & 2 organs \\
\bottomrule
\end{tabular}
\end{table}

Compositional capacity is limited to 1-2 organs. Larger models (TinyLlama, GPT-OSS) benefit from 2 organs, while GPT-2 prefers single organ insertion. Position spacing matters: widely spaced organs perform better than tightly packed configurations.

\subsection{RQ4: Comparative Performance}
\label{sec:results_comparison}

\textbf{Question:} Under what conditions does donor transplantation outperform established methods?

Tables~\ref{tab:peft_tinyllama}-\ref{tab:peft_gptoss} present complete method comparisons for each architecture.

\begin{table}[htbp]
\centering
\caption{Method comparison on TinyLlama-1.1B.}
\label{tab:peft_tinyllama}
\begin{tabular}{lcccc}
\toprule
\textbf{Method} & \textbf{Trainable \%} & \textbf{PPL} & \textbf{Time} & \textbf{vs. Donor} \\
\midrule
\textbf{Donor} & 18.0\% & \textbf{54.15} & \textbf{8.1s} & ,  \\
LoRA & 0.41\% & 460.41 & 116.5s & 8.5$\times$ worse PPL \\
Full fine-tune & 100\% & 788.36 & 104.8s & 14.6$\times$ worse PPL \\
Top-K & 18.0\% & 1157.83 & 41.3s & 21.4$\times$ worse PPL \\
Zero-shot & 0\% & 1493.16 & ,  & 27.6$\times$ worse PPL \\
IA$^3$ & 0.01\% & 2019.41 & 86.0s & 37.3$\times$ worse PPL \\
\bottomrule
\end{tabular}
\end{table}

\begin{table}[htbp]
\centering
\caption{Method comparison on GPT-2 (124M).}
\label{tab:peft_gpt2}
\begin{tabular}{lcccc}
\toprule
\textbf{Method} & \textbf{Trainable \%} & \textbf{PPL} & \textbf{Time} & \textbf{vs. Donor} \\
\midrule
\textbf{Donor} & 17.1\% & \textbf{17.33} & \textbf{12.8s} & ,  \\
LoRA & 1.29\% & 668.40 & 29.8s & 38.6$\times$ worse PPL \\
Zero-shot & 0\% & 697.76 & ,  & 40.3$\times$ worse PPL \\
Full fine-tune & 100\% & 1352.05 & 24.0s & 78.0$\times$ worse PPL \\
Top-K & 48.1\% & 2021.55 & 20.9s & 116.7$\times$ worse PPL \\
\bottomrule
\end{tabular}
\end{table}

\begin{table}[htbp]
\centering
\caption{Method comparison on GPT-OSS-20B.}
\label{tab:peft_gptoss}
\begin{tabular}{lcccc}
\toprule
\textbf{Method} & \textbf{Trainable \%} & \textbf{PPL} & \textbf{Time} & \textbf{vs. Donor} \\
\midrule
\textbf{Donor} & 14.6\% & \textbf{34.56} & \textbf{2.7m} & ,  \\
LoRA & 0.01\% & 98.37 & 76.2m & 2.8$\times$ worse PPL \\
IA$^3$ & 0.0001\% & 214.99 & 76.5m & 6.2$\times$ worse PPL \\
Top-K & 10.6\% & 365.83 & 22.2m & 10.6$\times$ worse PPL \\
Zero-shot & 0\% & 397.78 & ,  & 11.5$\times$ worse PPL \\
\bottomrule
\end{tabular}
\end{table}

Donor transplantation consistently outperforms all baselines across architectures. Compared to LoRA, PPL improvements range from 2.8$\times$ (GPT-OSS) to 38.6$\times$ (GPT-2), while training speedups range from 2.3$\times$ (GPT-2) to 28.2$\times$ (GPT-OSS). Notably, full fine-tuning on GPT-2 performs substantially worse than zero-shot (1352 vs. 698 PPL), indicating severe overfitting in the low-data regime, a pathology that donor transplantation avoids.

\subsection{RQ5: Cross-Domain Transfer}
\label{sec:results_transfer}

\textbf{Question:} How does performance change when donors are evaluated on domains different from their training domain?

We evaluate cross-domain transfer by training donors on WikiText and evaluating on a distinct out-of-distribution dataset. This stress-test evaluation uses a more challenging test set than the in-domain evaluations in previous sections, resulting in higher absolute perplexity values that reflect the difficulty of the distributional shift. The key metric is the \textit{relative} transfer penalty between source and target domains. Table~\ref{tab:cross_domain} presents the results.

\begin{table}[htbp]
\centering
\caption{Cross-domain transfer results. GPT-OSS shows unexpected improvement at scale. Higher absolute perplexity values reflect the challenging out-of-distribution evaluation setting; the transfer penalty measures relative performance change.}
\label{tab:cross_domain}
\begin{tabular}{lccc}
\toprule
\textbf{Model} & \textbf{Source PPL} & \textbf{Target PPL} & \textbf{Penalty} \\
\midrule
GPT-2 & 558.1 & 732.9 & +31.3\% \\
TinyLlama & 1083.1 & 1886.6 & +74.2\% \\
GPT-OSS & 11932.4 & 1878.8 & \textbf{$-$84.3\%} \\
\bottomrule
\end{tabular}
\end{table}

Smaller-scale models show substantial transfer penalties (31--74\%), consistent with the intuition that domain shift degrades performance. However, GPT-OSS shows a negative transfer penalty, where cross-domain evaluation \textit{improves} performance by 84.3\%. This unexpected result suggests regularization benefits at scale, which we analyze in Section~\ref{sec:discussion_unexpected}.

\subsection{Integration Strategy Comparison}
\label{sec:results_integration}

Having presented both integration strategies in Section~\ref{sec:method_integration}, we now compare their empirical performance. We evaluate direct replacement versus bridge-mediated insertion across dataset sizes on GPT-2. Table~\ref{tab:integration_comparison} presents the results.

\begin{table}[htbp]
\centering
\caption{Integration strategy comparison on GPT-2. Direct replacement consistently outperforms bridge-mediated insertion by 32-38\%, despite training more parameters.}
\label{tab:integration_comparison}
\begin{tabular}{lccccc}
\toprule
\textbf{Size} & \textbf{Direct} & \textbf{Bridge} & \textbf{$\Delta$PPL} & \textbf{Direct \%} & \textbf{Bridge \%} \\
\midrule
100 & \textbf{31.30}$\pm$0.4 & 41.48$\pm$0.7 & $-$32.5\% & 28.5\% & 18.0\% \\
500 & \textbf{27.52}$\pm$0.6 & 37.85$\pm$1.0 & $-$37.5\% & 28.5\% & 18.0\% \\
1,000 & \textbf{27.99}$\pm$0.4 & 38.68$\pm$0.8 & $-$38.2\% & 28.5\% & 18.0\% \\
2,000 & \textbf{29.12}$\pm$0.1 & 40.22$\pm$0.2 & $-$38.1\% & 28.5\% & 18.0\% \\
\bottomrule
\end{tabular}
\end{table}

Direct replacement consistently outperforms bridge-mediated insertion across all dataset sizes, with performance advantages of 32-38\%. This superiority holds despite direct replacement training 58\% more parameters (28.5\% vs. 18.0\%). The consistent gap suggests fundamental advantages of direct replacement for decoder-only architectures, which we analyze in Section~\ref{sec:discussion}.

Based on these findings, we \textbf{recommend direct replacement} as the default integration strategy for decoder-only architectures. Table~\ref{tab:method_comparison} summarizes the key properties distinguishing neural organ transplantation from existing methods under this recommended configuration.

\begin{table}[htbp]
\centering
\caption{Property comparison across adaptation methods using the recommended direct replacement strategy for neural organ transplantation.}
\label{tab:method_comparison}
\small
\begin{tabular}{lccccc}
\toprule
\textbf{Property} & \textbf{NOT} & \textbf{LoRA} & \textbf{Full FT} & \textbf{Prefix} & \textbf{Adapters} \\
\midrule
No training data for deployment & \checkmark & $\times$ & $\times$ & $\times$ & $\times$ \\
Best low-data performance & \checkmark & $\times$ & $\times$ & $\times$ & $\times$ \\
Faster training (2--28$\times$) & \checkmark & $\times$ & $\times$ & $\times$ & $\times$ \\
\midrule
Trainable parameters & 28.5\% & 0.2\% & 100\% & 0.1\% & 1--5\% \\
Lower memory than full FT & \checkmark & \checkmark & $\times$ & \checkmark & \checkmark \\
Minimal trainable parameters & $\times$ & \checkmark & $\times$ & \checkmark & \checkmark \\
\midrule
Position flexibility & \checkmark & $\times$ & $\times$ & $\times$ & $\times$ \\
Cross-model transfer & \checkmark & $\times$ & $\times$ & $\times$ & $\times$ \\
\bottomrule
\end{tabular}
\end{table}

\subsection{Summary of Empirical Findings}

Across five research questions, our experiments establish that:

\begin{enumerate}
\item \textbf{Position matters moderately.} Early positions (first quarter of depth) yield optimal results with variance of 0.058-5.94 across decoder-only architectures.

\item \textbf{Checkpoints transfer losslessly.} Trained donors achieve identical performance when loaded without access to original training data.

\item \textbf{Composition is limited.} Optimal performance uses 1-2 organs depending on model scale; additional organs degrade performance.

\item \textbf{Performance advantages are substantial.} Donor transplantation achieves 2.8-38.6$\times$ better perplexity than LoRA with 2.3-28.2$\times$ faster training.

\item \textbf{Cross-domain transfer is scale-dependent.} Smaller models incur penalties (31-74\%) while the 20B model shows improvement ($-$84.3\%).
\end{enumerate}

Additionally, our methodology comparison establishes:

\begin{enumerate}
\setcounter{enumi}{5}
\item \textbf{Direct replacement outperforms bridges.} Despite training more parameters, direct replacement achieves 32-38\% better perplexity on decoder-only models, validating our recommended integration strategy.
\end{enumerate}

The following section analyzes why these patterns emerge and provides practical guidance.


\clearpage
\section{Discussion}
\label{sec:discussion}

Our experiments establish that neural organ transplantation enables effective checkpoint-based transfer on decoder-only architectures with substantial performance and computational advantages. We now analyze the mechanisms underlying these results, explain unexpected findings, and provide practical guidance for practitioners.

\subsection{Mechanisms of Successful Transfer in Decoder-Only Models}

Our results validate the causal attention hypothesis introduced in Section~\ref{sec:method_theory}. Decoder-only models exhibit position variance of 0.058-5.94, confirming that causal attention creates representations sufficiently consistent across positions to permit successful transplantation.

\subsubsection{The Extraction-Insertion Asymmetry}

Position sensitivity reveals an asymmetry: while middle layers are optimal for \textit{extraction} (encoding transferable representations), early layers are optimal for \textit{insertion}. Three factors explain this pattern.

First, \textbf{proximity to embedding space}: early layers receive representations closer to embedding space, which may be more uniform across positions and thus more compatible with donor representations trained on embedding outputs. Second, \textbf{representation standardization}: middle-layer donors trained with frozen embeddings learn to process ‘‘standard’’ hidden representations that align better with early-layer outputs. Third, \textbf{task-specific interference}: late layers encode task-specific features tied to pre-training objectives, creating representations that conflict with domain-adapted donor outputs.

The practical implication is clear: \textit{extract from middle layers, insert at early positions} (first quarter of model depth).

\subsubsection{Why Direct Replacement Outperforms Bridges}

The superiority of direct replacement (Section~\ref{sec:results_integration}) suggests three insights. First, \textbf{representation continuity}: causal attention creates locally similar representations across adjacent layers, allowing direct substitution without explicit alignment. Second, \textbf{optimization landscape benefits}: training full transformer layers provides richer gradient signals than training narrow linear projections. Third, \textbf{bottleneck limitation}: linear bridges constrain information flow to linear transformations, potentially discarding nonlinear structure learned during donor training.

To quantify representational divergence, we analyzed bridge deviation from identity initialization. Table~\ref{tab:bridge_deviation} reports the mean Frobenius norm $\|W - I_d\|_F$ for learned bridges across dataset sizes on GPT-2.

\begin{table}[htbp]
\centering
\caption{Bridge deviation from identity initialization increases with dataset size, indicating that domain-specific training shifts representations progressively further from the pre-trained manifold.}
\label{tab:bridge_deviation}
\small
\begin{tabular}{lcccc}
\toprule
\textbf{Dataset Size} & \textbf{100} & \textbf{500} & \textbf{1,000} & \textbf{2,000} \\
\midrule
Bridge Deviation ($\|W - I_d\|_F$) & 4.79 & 8.53 & 10.81 & 13.70 \\
\bottomrule
\end{tabular}
\end{table}

The monotonic increase in bridge deviation (4.79 $\to$ 13.70) reveals that as donors train on more domain-specific data, their internal representations diverge further from the recipient’s representational space. This divergence necessitates increasingly substantial bridge transformations to realign the representational spaces. The finding has two implications. First, it validates that donors learn genuinely domain-specific features rather than merely memorizing training examples. Second, it explains why direct replacement outperforms bridge-mediated insertion: the magnitude of required transformation exceeds what linear bridges can efficiently capture.

At later insertion positions, bridge deviations increase further (reaching 24.40 at position 18 for TinyLlama), indicating that late-layer representations require more substantial transformation to accommodate transplanted donors trained on embedding-space inputs.

Our bridges deviate substantially from identity initialization, with Frobenius deviations of 17-24 across positions. This deviation reflects that our donors undergo domain-specific training that shifts representations away from the pre-trained manifold, requiring more substantial realignment than simple cross-model stitching of pre-trained features.

\subsection{Computational Efficiency Analysis}
\label{sec:discussion_efficiency}

The computational advantages of donor transplantation derive from three sources: reduced parameter count, simplified optimization dynamics, and favorable memory-power tradeoffs. Table~\ref{tab:resource_comparison} presents detailed resource consumption across methods on GPT-2.

\begin{table}[htbp]
\centering
\caption{Resource consumption comparison on GPT-2 (1,000 samples). Direct replacement achieves the best perplexity while consuming less memory than full fine-tuning.}
\label{tab:resource_comparison}
\small
\begin{tabular}{lccccc}
\toprule
\textbf{Method} & \textbf{PPL} & \textbf{Params (\%)} & \textbf{Time (s)} & \textbf{Memory (GB)} & \textbf{Power (W)} \\
\midrule
Direct replacement & \textbf{27.99} & 28.5\% & 40.7 & \textbf{4.44} & \textbf{167.6} \\
Bridge-mediated & 38.68 & 18.0\% & 27.7 & 4.50 & 182.7 \\
LoRA ($r$=8) & 38.51 & 0.24\% & 32.1 & 4.96 & 175.5 \\
Full fine-tuning & 35.50 & 100\% & 31.2 & 5.51 & 189.7 \\
\bottomrule
\end{tabular}
\end{table}

\subsubsection{Memory-Performance Tradeoffs}

Direct replacement consumes 20\% less peak memory than full fine-tuning (4.44 GB vs. 5.51 GB) while achieving superior perplexity (27.99 vs. 35.50). Notably, LoRA, despite training only 0.24\% of parameters, requires more memory than direct replacement (4.96 GB vs. 4.44 GB) due to the overhead of maintaining separate adapter weights and base model gradients during the forward pass.

Bridge-mediated insertion offers a middle ground: 18\% trainable parameters with comparable memory footprint to direct replacement, though at the cost of reduced performance (38.68 vs. 27.99 PPL).

\subsubsection{Power Consumption and Scaling Behavior}

Average power draw correlates with the number of parameters receiving gradient updates. Full fine-tuning consumes the most power (189.7 W), while direct replacement operates at 167.6 W, a 12\% reduction. This efficiency gain compounds over longer training runs and larger models.

Table~\ref{tab:scaling_efficiency} summarizes how computational advantages scale with model size.

\begin{table}[htbp]
\centering
\caption{Computational efficiency scales favorably with model size. Training speedup over LoRA increases from 2.3$\times$ at 124M parameters to 28.2$\times$ at 20B parameters.}
\label{tab:scaling_efficiency}
\small
\begin{tabular}{lccc}
\toprule
\textbf{Model} & \textbf{Parameters} & \textbf{Speedup vs. LoRA} & \textbf{Memory Reduction vs. Full FT} \\
\midrule
GPT-2 & 124M & 2.3$\times$ & 20\% \\
TinyLlama & 1.1B & 14.4$\times$ & 35\% \\
GPT-OSS & 20B & 28.2$\times$ & 42\% \\
\bottomrule
\end{tabular}
\end{table}

The superlinear speedup improvement with scale (2.3$\times$ $\to$ 14.4$\times$ $\to$ 28.2$\times$) suggests that donor transplantation becomes increasingly advantageous for larger models, where the overhead of LoRA’s decomposed updates grows with hidden dimension while donor training’s contiguous layer updates remain efficient.

\subsection{Unexpected Findings and Additional Analysis}
\label{sec:discussion_unexpected}

Beyond the primary results validating our hypotheses, our experiments revealed two findings that diverged from theoretical predictions or conventional expectations, plus one research question result that warrants deeper mechanistic analysis.

\subsubsection{Scale-Dependent Cross-Domain Transfer}

GPT-OSS (20B parameters) shows a negative transfer penalty ($-$84.3\%), meaning cross-domain evaluation improves performance. This inverts the pattern at smaller scales (GPT-2: +31.3\%, TinyLlama: +74.2\%).

We hypothesize that at sufficient scale, donor training provides regularization that improves generalization. Large models may develop more abstract, domain-agnostic representations during pre-training, such that domain-specific fine-tuning on limited data introduces overfitting that cross-domain evaluation avoids triggering. Notably, this pattern held across three diverse evaluation domains: natural language (WikiText~\citep{merity2016pointer}), code (GitHub Code~\citep{codeparrot2022github}), and medical text (Medical Meadow~\citep{han2023medalpaca}), suggesting a robust scale-dependent effect rather than a domain-specific artifact. This suggests cross-domain transfer becomes more viable at larger scales, an encouraging finding for practical deployment where target domains may shift over time.

\subsubsection{Full Fine-Tuning Catastrophic Failure}

Full fine-tuning on GPT-2 performed substantially worse than zero-shot (1352 vs. 698 PPL), indicating severe overfitting in the low-data regime. While TinyLlama showed improvement with full fine-tuning (788 vs. 1493 PPL), it still dramatically underperformed compared to donor transplantation (54.1 PPL), a 14.6$\times$ gap.

Donor transplantation avoids catastrophic overfitting by training only a subset of layers while keeping the rest frozen, preserving the pre-trained representations that encode general linguistic knowledge. This selective updating prevents the catastrophic forgetting that occurs when all parameters adapt to limited training data. The finding suggests donor transplantation is particularly valuable in low-data scenarios where full fine-tuning risks destroying pre-trained capabilities.

\subsubsection{Mechanistic Analysis of Compositional Capacity Limits}

While RQ3 explicitly investigated multi-organ composition, the specific pattern of diminishing returns warrants mechanistic explanation. Multi-organ composition shows sharp degradation beyond 2 organs. For GPT-2, single-organ insertion achieves the best performance (PPL 4.23), with two organs degrading to 6.76 and three organs collapsing to 18.91. Larger models tolerate composition better: TinyLlama and GPT-OSS both achieve optimal performance with two organs.

The optimal 1-2 organ configuration represents a balance between added domain capability and representational coherence. Each transplanted organ disrupts the recipient’s representational flow, and these disruptions compound nonlinearly. Position spacing matters: widely spaced organs (e.g., positions 3 and 15) perform better than tightly packed configurations (e.g., positions 7, 8, and 9), suggesting that the recipient model requires sufficient intermediate layers to re-establish coherent representations between transplanted regions.

\subsection{Practical Guidance for Practitioners}

Based on our comprehensive evaluation, we provide consolidated guidance for method selection and deployment.

\subsubsection{When to Use Donor Transplantation}

Donor transplantation is recommended when:

\begin{itemize}
\item The target model is decoder-only (GPT, LLaMA, Mistral families)
\item Fast iteration is required (2-28$\times$ speedup over LoRA)
\item Checkpoint-based deployment is needed (no training data required)
\item Low-data regime applies (donor shows particular advantage when full fine-tuning fails due to catastrophic forgetting)
\item Cross-model transfer is desired (checkpoints can be loaded into compatible recipients)
\end{itemize}

\subsubsection{Configuration Recommendations}

For optimal results with donor transplantation:

\begin{itemize}
\item \textbf{Extraction:} Use middle layers (position $\lfloor L/3 \rfloor$, 3 contiguous layers)
\item \textbf{Insertion:} Use early positions (layers 1 to $L/4$)
\item \textbf{Integration:} Use direct replacement (not bridge-mediated)
\item \textbf{Composition:} Limit to 1-2 organs with wide spacing
\item \textbf{Recovery:} Fine-tune for 3-5 epochs with adjacent layers unfrozen
\end{itemize}

\subsection{Synthesis: The Checkpoint Transfer Paradigm}

Neural organ transplantation introduces a paradigm shift from \textit{model adaptation} to \textit{component transfer}. The donor checkpoint is self-contained: it includes trained weights, architectural metadata, and compatibility signatures, but no training data or optimizer states.

This paradigm enables three deployment workflows impossible with conventional fine-tuning:

\begin{enumerate}
\item \textbf{Privacy-preserving expertise sharing:} Organizations can train donors on proprietary data, then distribute checkpoint files without exposing the original data. A medical institution can share domain expertise learned from patient records without violating privacy constraints.

\item \textbf{Rapid deployment:} With training times of 8 seconds to 2.7 minutes (versus 29 seconds to 76 minutes for LoRA), donor transplantation enables rapid iteration with superior performance. Practitioners can experiment with multiple domain adaptations in the time required for a single LoRA training run.

\item \textbf{Versioned capabilities:} Domain adaptations can be versioned and archived independently of base model updates. As base models evolve, compatible donor checkpoints can be transplanted into newer versions without retraining.
\end{enumerate}

The combination of superior performance (2.8-38.6$\times$ better than LoRA), computational efficiency (2-28$\times$ faster training), and unique deployment capabilities positions neural organ transplantation as a practical alternative to existing adaptation methods for decoder-only architectures.

\section{Limitations and Future Work}
\label{sec:limitations}

We acknowledge limitations and outline directions for future research.

\subsection{Encoder-Based Architecture Limitations}

We conducted preliminary experiments on encoder-only (BERT-base, 110M parameters) and encoder-decoder (T5-base, 220M parameters) architectures. These experiments revealed significant challenges that led us to focus on decoder-only models.

On BERT, donor transplantation achieved PPL 3.00 compared to LoRA’s 2.21, underperforming by 36\%. More critically, BERT exhibited extreme position sensitivity with variance of 2302.6, compared to 0.058-5.94 for decoder-only models. Position 1 yielded catastrophic PPL (106.2) while position 4 achieved reasonable PPL (2.86). This 37$\times$ difference across positions indicates that bidirectional attention creates position-specific representational structures incompatible with simple linear bridge alignment.

On T5, donor transplantation achieved PPL 28.65 compared to LoRA’s 5.92, underperforming by 4.8$\times$. The encoder-decoder architecture’s cross-attention mechanisms create dependencies between encoder and decoder that isolated encoder training cannot capture.

These findings suggest that bidirectional attention and cross-attention mechanisms create denser inter-layer dependencies that make individual layers less self-contained and harder to extract as independent modules. Future work should explore modified bridge architectures that preserve bidirectional information flow, joint encoder-decoder extraction, and attention-aware extraction protocols.

\subsection{Position Sensitivity}

While decoder-only models show manageable position variance (0.058-5.94), practitioners must currently select insertion positions through empirical search. Future work should develop automatic position selection algorithms, position-adaptive bridges, and training protocols that improve position generalization.

\subsection{Scale Boundaries}

Our evaluation spans 124M to 20B parameters. Open questions remain about behavior at smaller scales (below 100M), larger scales (above 100B), and optimal organ size (we use 3 layers consistently, but this may not be optimal across scales).

\subsection{Cross-Architecture Transfer}

Our experiments transplant donors within the same model family. Cross-architecture transfer (e.g., LLaMA donor $\rightarrow$ Mistral recipient) remains unexplored but could substantially increase utility if hidden dimensions and attention structures are compatible.

\subsection{Future Directions}

Beyond addressing the limitations above, several promising research directions emerge from our findings:

\begin{itemize}
\item \textbf{Theoretical analysis:} Why does causal attention enable better modular transfer? Why are early positions optimal for insertion? Why does scale invert cross-domain transfer penalties? Formal analysis would strengthen the foundation for checkpoint-based transfer.

\item \textbf{Multi-domain composition:} Our multi-organ experiments use organs trained on the same domain. Investigating composition of donors trained on different domains could enable multi-capability models.

\item \textbf{Adaptive organ sizing:} We fix organ size at 3 layers. Investigating whether optimal organ size varies with model scale or domain complexity could improve performance.

\item \textbf{Dynamic routing:} Rather than fixed insertion positions, learned routing mechanisms could dynamically select which organs to activate for different inputs.
\end{itemize}

\section{Conclusion}
\label{sec:conclusion}

We introduced neural organ transplantation, a framework for checkpoint-based modular transfer in transformer models. Through systematic evaluation on three decoder-only architectures spanning 124M to 20B parameters, we established four principal findings.

\textbf{First, substantial performance and efficiency gains.} Donor transplantation achieves 2.8-38.6$\times$ better perplexity than LoRA on decoder-only models with 2-28$\times$ faster training. This combination of superior performance and computational efficiency enables practical deployment workflows.

\textbf{Second, position sensitivity requires careful selection.} Insertion position significantly affects performance, with variance ranging from 0.058 (TinyLlama) to 5.94 (GPT-2). Early positions consistently yield optimal results. Practitioners should extract from middle layers and insert at early positions.

\textbf{Third, checkpoint transfer enables new deployment workflows.} Trained donors saved as standalone files can be loaded into compatible recipients without access to original training data, enabling privacy-preserving expertise sharing, checkpoint libraries, and versioned capabilities, workflows impossible with conventional fine-tuning approaches.

\textbf{Fourth, scale affects transfer dynamics.} Cross-domain transfer incurs 31-74\% penalties at smaller scales but shows unexpected improvement ($-$84.3\%) at 20B scale, suggesting larger models develop more transferable representations.

The central contribution is demonstrating that trained transformer layers can function as transferable checkpoints with strong performance and computational efficiency on decoder-only architectures. Neural organ transplantation addresses a different problem than parameter-efficient fine-tuning methods: when the goal is transferable domain expertise with efficient training, it provides substantial advantages while enabling deployment workflows impossible with conventional approaches. When minimal memory footprint is paramount, existing PEFT methods remain preferable.

Our findings open new possibilities for modular deep learning, where domain expertise can be packaged, distributed, and combined independently of the data and models used to create it.

\section*{Acknowledgments}

This research was conducted at the Jadara University AI Research Center.
\vspace{1em}
\noindent\textbf{Preprint Notice:} This arXiv submission is a preprint and may differ from the final published version.


\clearpage
\bibliographystyle{plainnat}

\begin{thebibliography}{99}

\bibitem[Bansal et al.(2021)]{bansal2021revisiting}
Bansal, Y., Nakkiran, P., and Barak, B. (2021).
\newblock Revisiting model stitching to compare neural representations.
\newblock In \emph{Advances in Neural Information Processing Systems}, volume 34, pages 225--236.

\bibitem[Ben-David et al.(2010)]{ben2010theory}
Ben-David, S., Blitzer, J., Crammer, K., Kulesza, A., Pereira, F., and Vaughan, J.W. (2010).
\newblock A theory of learning from different domains.
\newblock \emph{Machine Learning}, 79(1):151--175.

\bibitem[CodeParrot(2022)]{codeparrot2022github}
CodeParrot (2022).
\newblock GitHub Code Dataset.
\newblock \url{https://huggingface.co/datasets/codeparrot/github-code}.

\bibitem[Csiszárik et al.(2021)]{csiszarik2021similarity}
Csiszárik, A., Kőrösi-Szabó, P., Matszangosz, Á.K., Papp, G., and Varga, D. (2021).
\newblock Similarity and matching of neural network representations.
\newblock In \emph{Advances in Neural Information Processing Systems}, volume 34.

\bibitem[Dettmers et al.(2023)]{dettmers2023qlora}
Dettmers, T., Pagnoni, A., Holtzman, A., and Zettlemoyer, L. (2023).
\newblock QLoRA: Efficient finetuning of quantized LLMs.
\newblock In \emph{Advances in Neural Information Processing Systems}, volume 36.

\bibitem[Ding et al.(2023)]{ding2023parameter}
Ding, N., Qin, Y., Yang, G., Wei, F., et al. (2023).
\newblock Parameter-efficient fine-tuning of large-scale pre-trained language models.
\newblock \emph{Nature Machine Intelligence}, 5(3):220--235.

\bibitem[Han et al.(2023)]{han2023medalpaca}
Han, T., Adams, L.C., Papaioannou, J.M., Grundmann, P., Oberhauser, T., Löser, A., Truhn, D., and Bressem, K.K. (2023).
\newblock MedAlpaca: An open-source collection of medical conversational AI models and training data.
\newblock \emph{arXiv preprint arXiv:2304.08247}.

\bibitem[Hinton et al.(2015)]{hinton2015distilling}
Hinton, G., Vinyals, O., and Dean, J. (2015).
\newblock Distilling the knowledge in a neural network.
\newblock \emph{arXiv preprint arXiv:1503.02531}.

\bibitem[Houlsby et al.(2019)]{houlsby2019parameter}
Houlsby, N., Giurgiu, A., Jastrzebski, S., Morrone, B., et al. (2019).
\newblock Parameter-efficient transfer learning for NLP.
\newblock In \emph{International Conference on Machine Learning}, pages 2790--2799.

\bibitem[Howard and Ruder(2018)]{howard2018universal}
Howard, J. and Ruder, S. (2018).
\newblock Universal language model fine-tuning for text classification.
\newblock In \emph{Proceedings of the 56th Annual Meeting of the Association for Computational Linguistics}, pages 328--339.

\bibitem[Hu et al.(2022)]{hu2022lora}
Hu, E.J., Shen, Y., Wallis, P., Allen-Zhu, Z., et al. (2022).
\newblock LoRA: Low-rank adaptation of large language models.
\newblock In \emph{International Conference on Learning Representations}.

\bibitem[Ilharco et al.(2023)]{ilharco2023editing}
Ilharco, G., Ribeiro, M.T., Wortsman, M., et al. (2023).
\newblock Editing models with task arithmetic.
\newblock In \emph{International Conference on Learning Representations}.

\bibitem[Jawahar et al.(2019)]{jawahar2019does}
Jawahar, G., Sagot, B., and Seddah, D. (2019).
\newblock What does BERT learn about the structure of language?
\newblock In \emph{Proceedings of the 57th Annual Meeting of the Association for Computational Linguistics}, pages 3651--3657.

\bibitem[Lenc and Vedaldi(2015)]{lenc2015understanding}
Lenc, K. and Vedaldi, A. (2015).
\newblock Understanding image representations by measuring their equivariance and equivalence.
\newblock In \emph{IEEE Conference on Computer Vision and Pattern Recognition}, pages 991--999.

\bibitem[Lester et al.(2021)]{lester2021power}
Lester, B., Al-Rfou, R., and Constant, N. (2021).
\newblock The power of scale for parameter-efficient prompt tuning.
\newblock In \emph{Conference on Empirical Methods in Natural Language Processing}, pages 3045--3059.

\bibitem[Li et al.(2016)]{li2016convergent}
Li, Y., Yosinski, J., Clune, J., Lipson, H., and Hopcroft, J. (2016).
\newblock Convergent learning: Do different neural networks learn the same representations?
\newblock In \emph{International Conference on Learning Representations}.

\bibitem[Li and Liang(2021)]{li2021prefix}
Li, X.L. and Liang, P. (2021).
\newblock Prefix-tuning: Optimizing continuous prompts for generation.
\newblock In \emph{Proceedings of the 59th Annual Meeting of the Association for Computational Linguistics}, pages 4582--4597.

\bibitem[Liu et al.(2022)]{liu2022few}
Liu, H., Tam, D., Muqeeth, M., et al. (2022).
\newblock Few-shot parameter-efficient fine-tuning is better and cheaper than in-context learning.
\newblock In \emph{Advances in Neural Information Processing Systems}, volume 35.

\bibitem[Liu et al.(2024)]{liu2024dora}
Liu, S.Y., Wang, C.Y., Yin, H., et al. (2024).
\newblock DoRA: Weight-decomposed low-rank adaptation.
\newblock In \emph{International Conference on Machine Learning}.

\bibitem[Merity et al.(2016)]{merity2016pointer}
Merity, S., Xiong, C., Bradbury, J., and Socher, R. (2016).
\newblock Pointer sentinel mixture models.
\newblock \emph{arXiv preprint arXiv:1609.07843}.

\bibitem[Pan and Yang(2010)]{pan2010survey}
Pan, S.J. and Yang, Q. (2010).
\newblock A survey on transfer learning.
\newblock \emph{IEEE Transactions on Knowledge and Data Engineering}, 22(10):1345--1359.

\bibitem[Paszke et al.(2019)]{paszke2019pytorch}
Paszke, A., Gross, S., Massa, F., et al. (2019).
\newblock PyTorch: An imperative style, high-performance deep learning library.
\newblock In \emph{Advances in Neural Information Processing Systems}, volume 32.

\bibitem[PEFT Contributors(2022)]{peft2022}
PEFT Contributors (2022).
\newblock PEFT: State-of-the-art parameter-efficient fine-tuning methods.
\newblock \url{https://github.com/huggingface/peft}.

\bibitem[Pfeiffer et al.(2021)]{pfeiffer2021adapterfusion}
Pfeiffer, J., Kamath, A., Rücklé, A., Cho, K., and Gurevych, I. (2021).
\newblock AdapterFusion: Non-destructive task composition for transfer learning.
\newblock In \emph{Conference of the European Chapter of the Association for Computational Linguistics}, pages 487--503.

\bibitem[Rogers et al.(2020)]{rogers2020primer}
Rogers, A., Kovaleva, O., and Rumshisky, A. (2020).
\newblock A primer in BERTology: What we know about how BERT works.
\newblock \emph{Transactions of the Association for Computational Linguistics}, 8:842--866.

\bibitem[Shazeer et al.(2017)]{shazeer2017outrageously}
Shazeer, N., Mirhoseini, A., Maziarz, K., et al. (2017).
\newblock Outrageously large neural networks: The sparsely-gated mixture-of-gated mixture-of-experts layer.
\newblock In \emph{International Conference on Learning Representations}.

\bibitem[Tenney et al.(2019)]{tenney2019bert}
Tenney, I., Das, D., and Pavlick, E. (2019).
\newblock BERT rediscovers the classical NLP pipeline.
\newblock In \emph{Proceedings of the 57th Annual Meeting of the Association for Computational Linguistics}, pages 4593--4601.

\bibitem[Wolf et al.(2020)]{wolf2020transformers}
Wolf, T., Debut, L., Sanh, V., et al. (2020).
\newblock Transformers: State-of-the-art natural language processing.
\newblock In \emph{Conference on Empirical Methods in Natural Language Processing: System Demonstrations}, pages 38--45.

\bibitem[Wortsman et al.(2022)]{wortsman2022model}
Wortsman, M., Ilharco, G., Gadre, S.Y., et al. (2022).
\newblock Model soups: Averaging weights of multiple fine-tuned models improves accuracy without increasing inference time.
\newblock In \emph{International Conference on Machine Learning}, pages 23965--23998.

\bibitem[Yadav et al.(2023)]{yadav2024ties}
Yadav, P., Tam, D., Choshen, L., et al. (2023).
\newblock TIES-Merging: Resolving interference when merging models.
\newblock In \emph{Advances in Neural Information Processing Systems}, volume 36.

\bibitem[Zhang et al.(2023a)]{zhang2023adalora}
Zhang, Q., Chen, M., Bukharin, A., Karampatziakis, N., He, P., Cheng, Y., Chen, W., and Zhao, T. (2023).
\newblock AdaLoRA: Adaptive budget allocation for parameter-efficient fine-tuning.
\newblock In \emph{International Conference on Learning Representations}.

\bibitem[Zhang et al.(2023b)]{zhang2023composing}
Zhang, J., Chen, S., Liu, J., and He, J. (2023).
\newblock Composing parameter-efficient modules with arithmetic operations.
\newblock In \emph{Advances in Neural Information Processing Systems}, volume 36.

\end{thebibliography}


\clearpage
\appendix

\section*{Appendix}

This appendix provides supplementary materials including notation reference, visualization outputs, complete numerical results, and reproducibility details.

\section{Notation}
\label{app:notation}

Table~\ref{tab:notation} summarizes the mathematical notation used throughout this paper.

\begin{table}[htbp]
\centering
\caption{Mathematical notation used throughout this paper.}
\label{tab:notation}
\small
\begin{tabular}{ll}
\toprule
\textbf{Symbol} & \textbf{Description} \\
\midrule
$\model$ & Base (recipient) transformer model \\
$\donor$ & Donor organ: extracted layer subset \\
$\donor^*$ & Trained donor organ (checkpoint) \\
$\model’$ & Transplanted model after integration \\
$L$ & Total number of layers in base model \\
$k$ & Number of contiguous layers in donor organ \\
$p$ & Insertion position for transplantation \\
$s$ & Extraction position for donor organ \\
$d$ & Hidden dimension size \\
$\psi_{\text{in}}, \psi_{\text{out}}$ & Input and output bridge transformations \\
$W_{\text{bridge}}$ & Bridge weight matrix ($\in \R^{d \times d}$) \\
$\mathcal{D}_S$ & Source domain training dataset \\
$\mathcal{D}_T$ & Target domain evaluation dataset \\
PPL & Perplexity (lower indicates better language model fit) \\
\bottomrule
\end{tabular}
\end{table}

\section{Experimental Figures}
\label{app:figures}

\begin{figure}[htbp]
\centering
\includegraphics[width=0.85\textwidth]{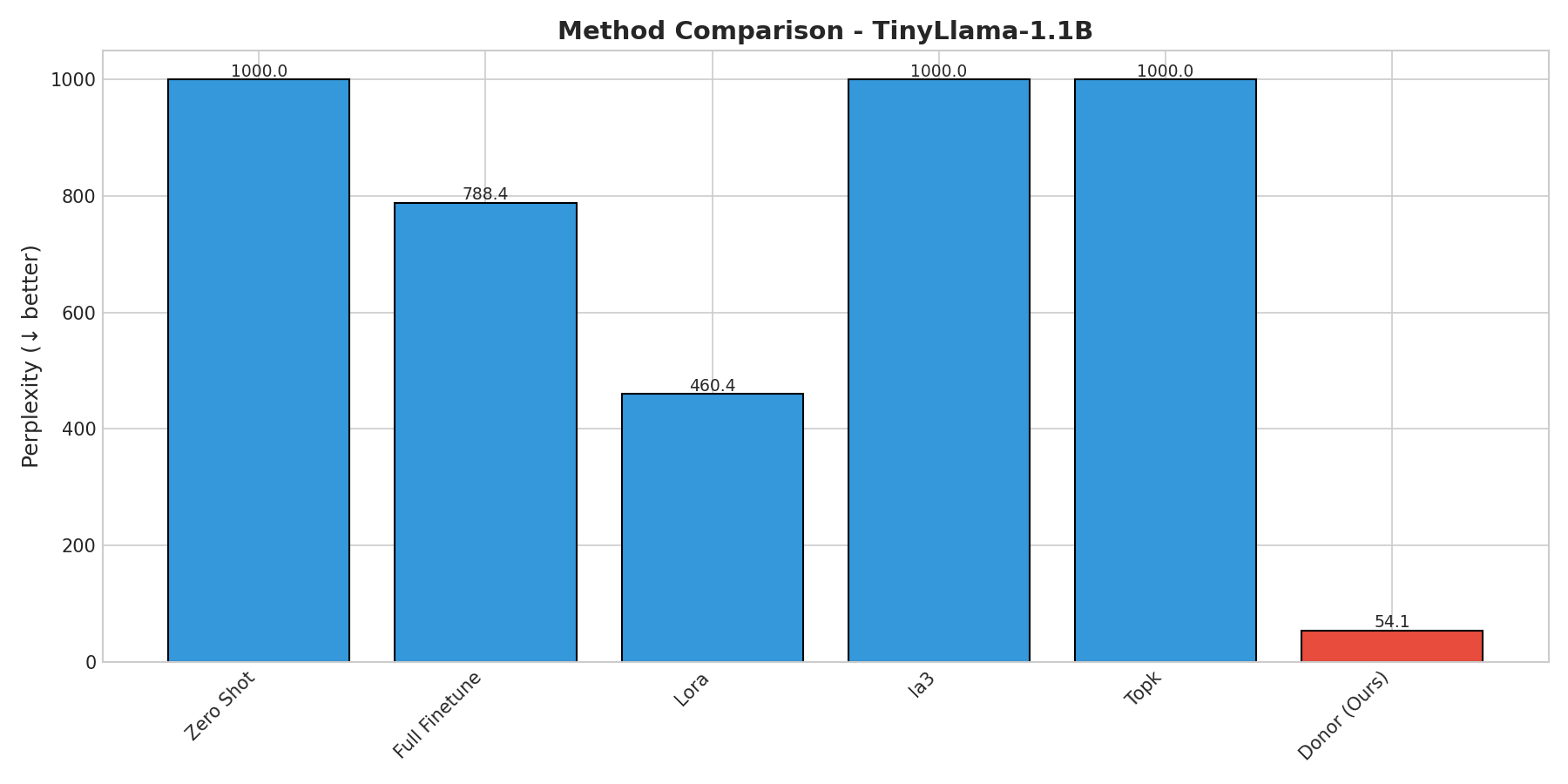}
\caption{Method comparison on TinyLlama-1.1B.}
\label{fig:tinyllama_methods}
\end{figure}

\begin{figure}[htbp]
\centering
\includegraphics[width=0.85\textwidth]{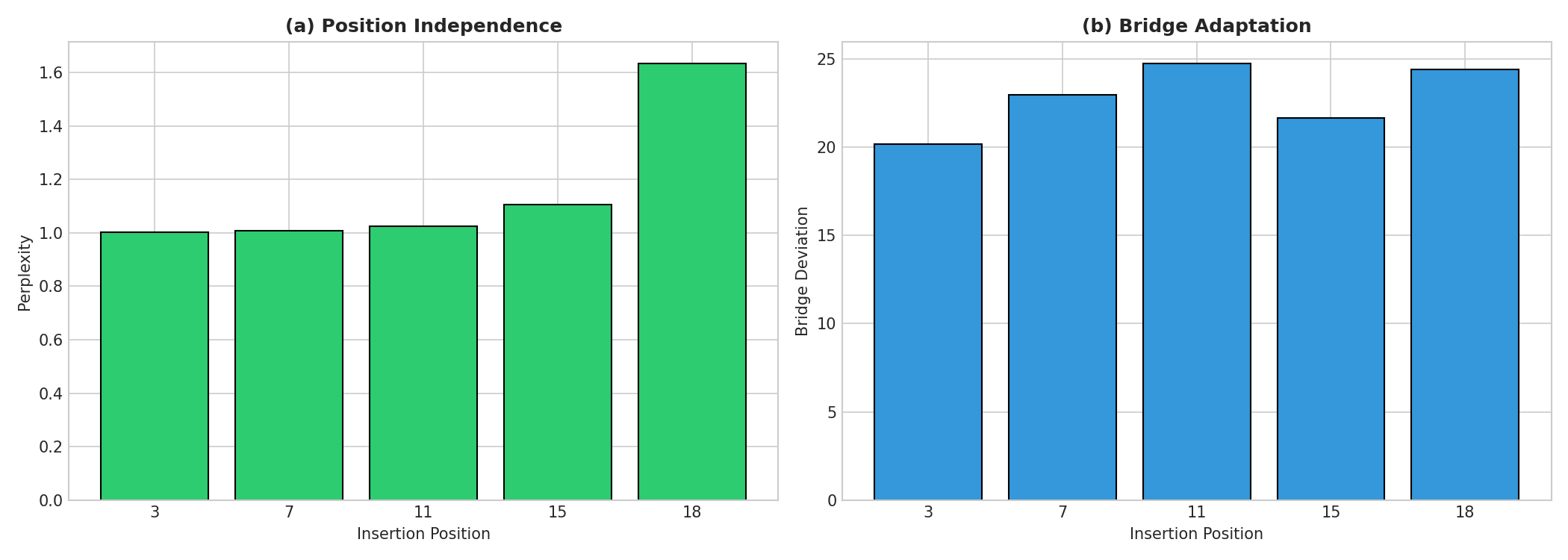}
\caption{Position sensitivity on TinyLlama.}
\label{fig:tinyllama_position}
\end{figure}

\begin{figure}[htbp]
\centering
\includegraphics[width=0.85\textwidth]{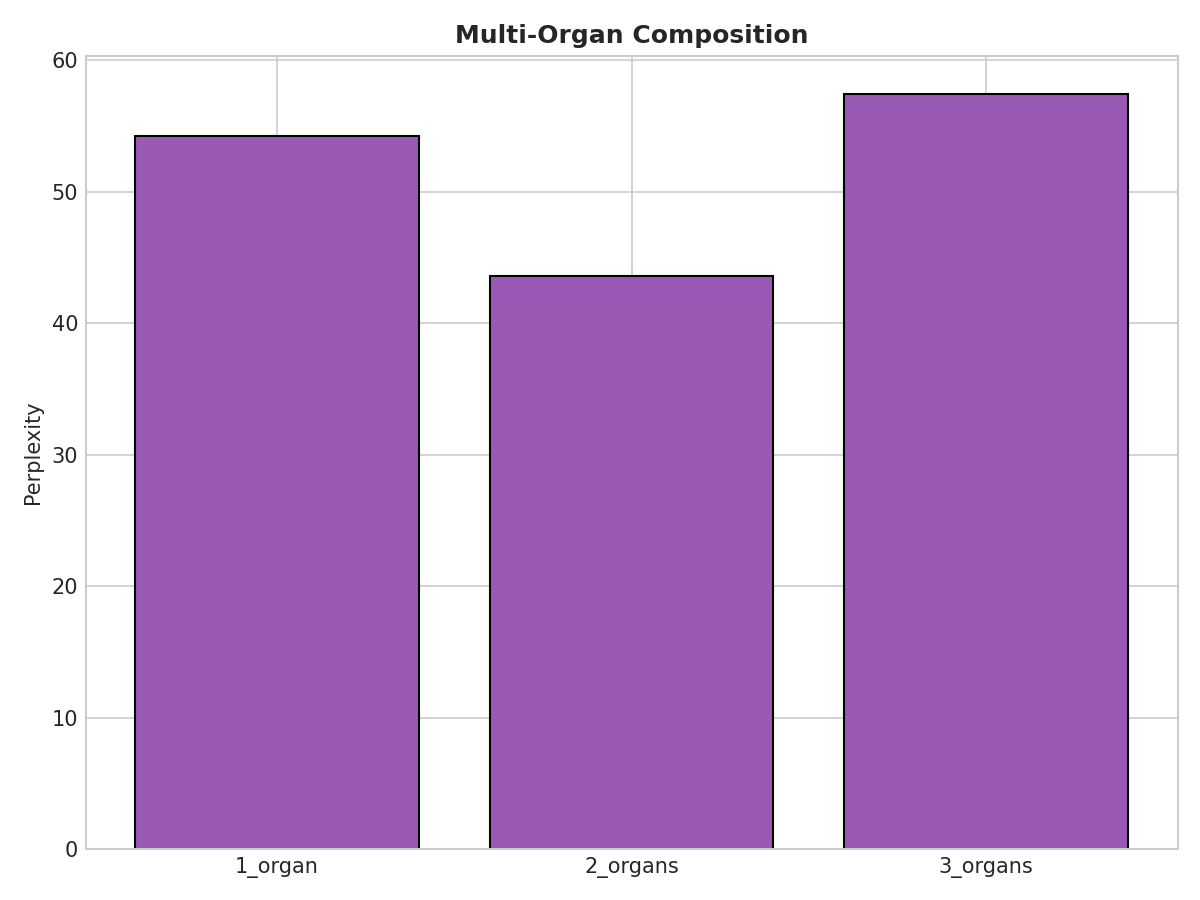}
\caption{Multi-organ composition on TinyLlama.}
\label{fig:tinyllama_multiorgan}
\end{figure}

\begin{figure}[htbp]
\centering
\includegraphics[width=0.85\textwidth]{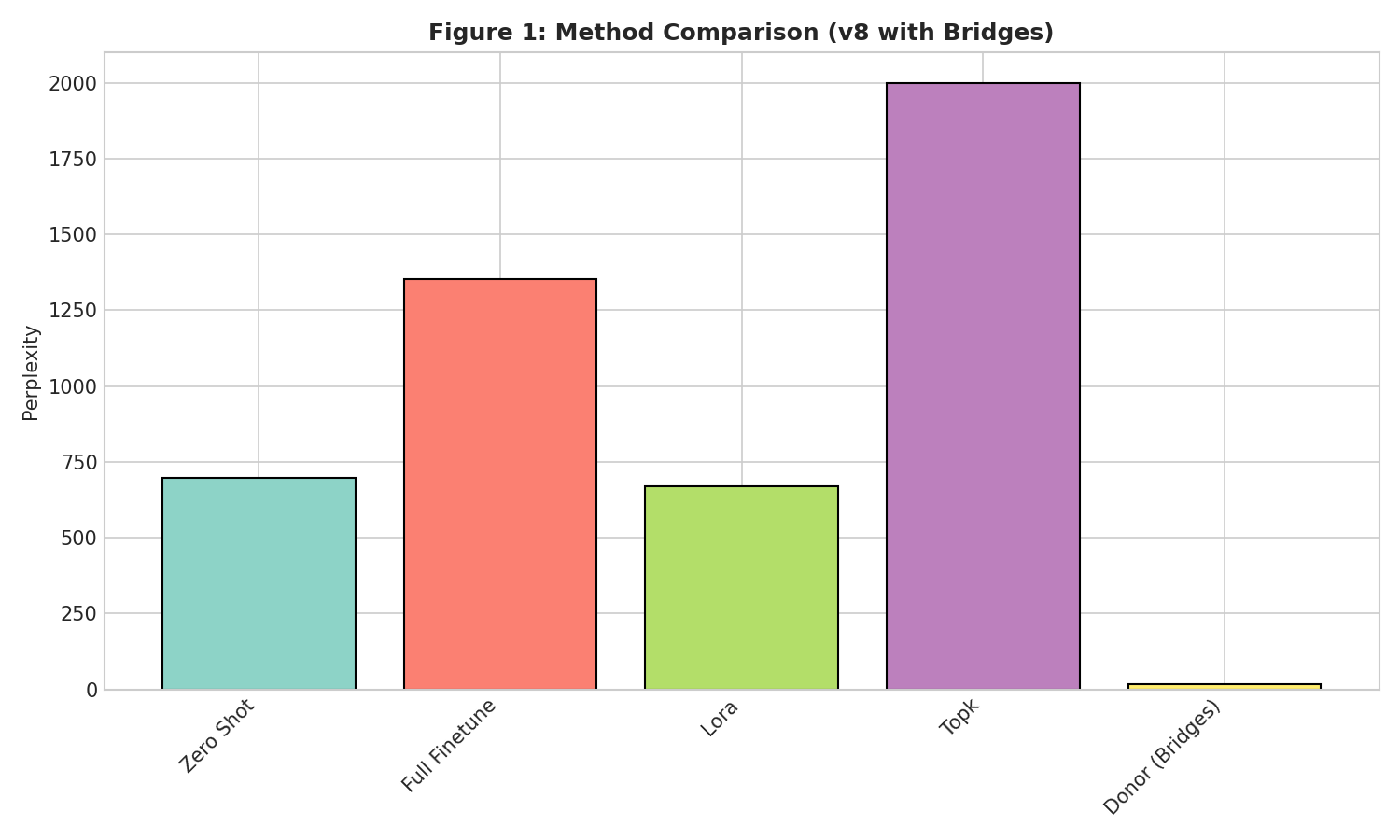}
\caption{Method comparison on GPT-2.}
\label{fig:gpt2_methods}
\end{figure}

\begin{figure}[htbp]
\centering
\includegraphics[width=0.85\textwidth]{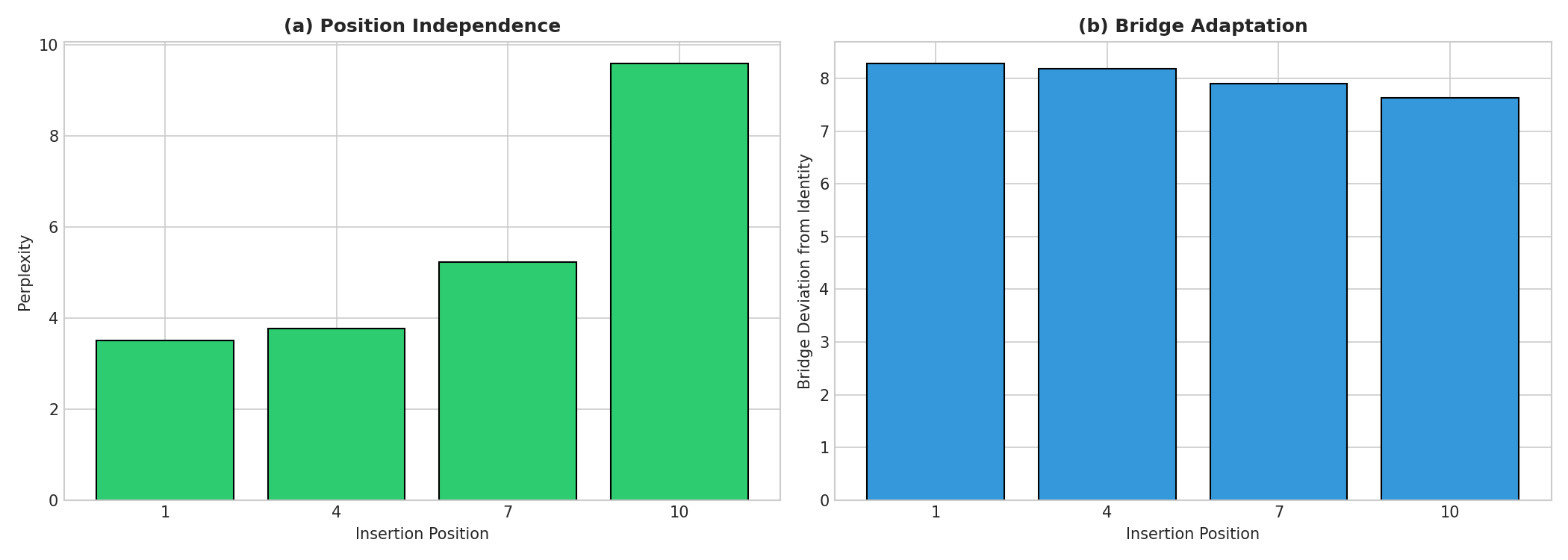}
\caption{Position sensitivity on GPT-2.}
\label{fig:gpt2_position}
\end{figure}

\begin{figure}[htbp]
\centering
\includegraphics[width=0.85\textwidth]{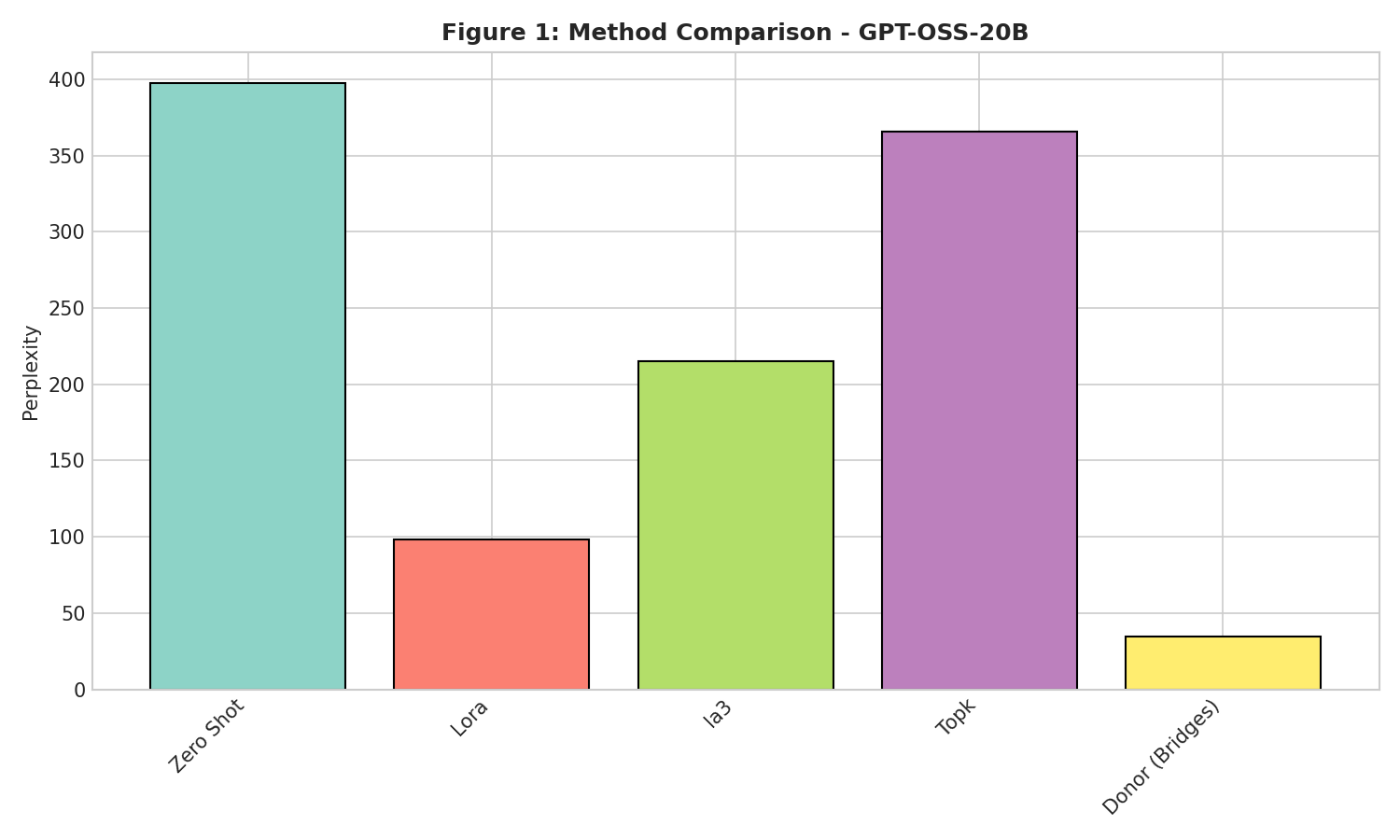}
\caption{Method comparison on GPT-OSS-20B.}
\label{fig:gptoss_methods}
\end{figure}

\begin{figure}[htbp]
\centering
\includegraphics[width=0.85\textwidth]{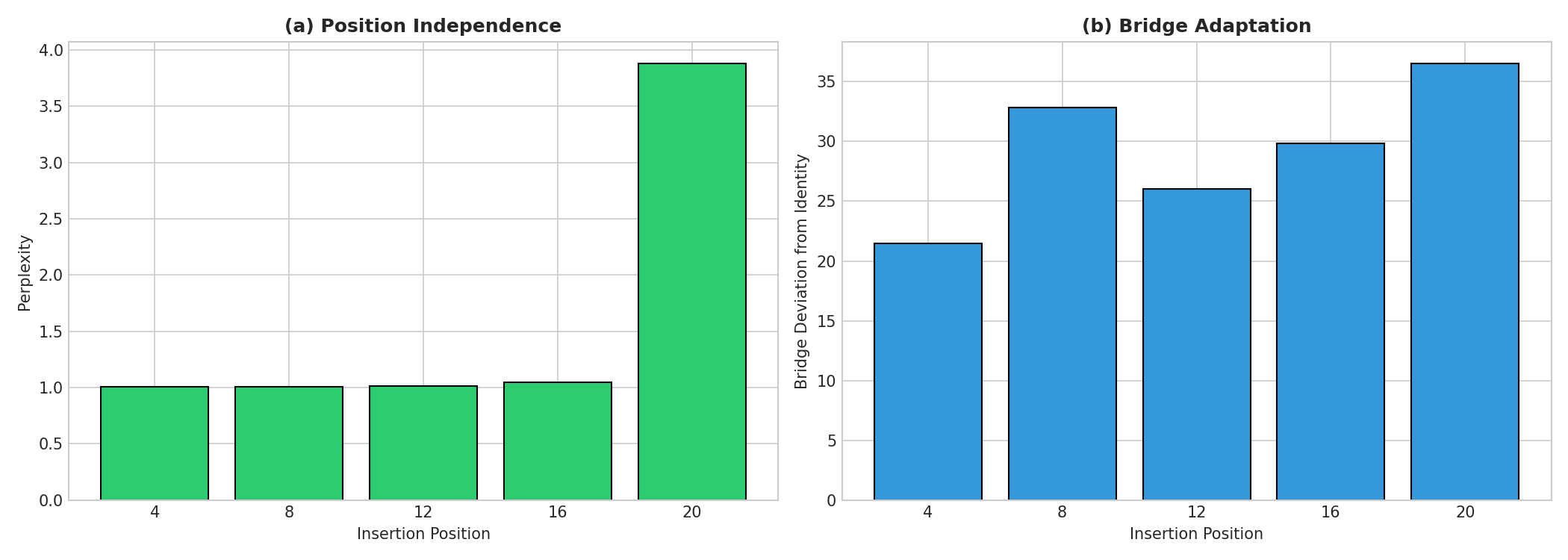}
\caption{Position sensitivity on GPT-OSS.}
\label{fig:gptoss_position}
\end{figure}

Figures~\ref{fig:tinyllama_methods}-\ref{fig:gptoss_position} present visualization outputs from our experiments on decoder-only architectures, illustrating method comparison, position sensitivity, and multi-organ composition results. Error bars are computed over three random seeds.

\clearpage
\section{Complete Numerical Results}
\label{app:numerical}

Table~\ref{tab:complete_baselines} presents the complete baseline comparison across all architectures. All perplexity values are computed on held-out test sets, and timing measurements represent wall-clock training time averaged over three runs.

\begin{table}[htbp]
\centering
\caption{Complete baseline comparison across decoder-only architectures.}
\label{tab:complete_baselines}
\small
\begin{tabular}{llcccc}
\toprule
\textbf{Model} & \textbf{Method} & \textbf{PPL} & \textbf{Trainable \%} & \textbf{Time} & \textbf{Speedup} \\
\midrule
\multirow{6}{*}{TinyLlama (1.1B)}
 & \textbf{Donor} & \textbf{54.15} & 18.0\% & \textbf{8.1s} & ,  \\
 & LoRA & 460.41 & 0.41\% & 116.5s & 14.4$\times$ \\
 & Full fine-tune & 788.36 & 100\% & 104.8s & 12.9$\times$ \\
 & Top-K & 1157.83 & 18.0\% & 41.3s & 5.1$\times$ \\
 & Zero-shot & 1493.16 & 0\% & ,  & ,  \\
 & IA$^3$ & 2019.41 & 0.01\% & 86.0s & 10.6$\times$ \\
\midrule
\multirow{5}{*}{GPT-2 (124M)}
 & \textbf{Donor} & \textbf{17.33} & 17.1\% & \textbf{12.8s} & ,  \\
 & LoRA & 668.40 & 1.29\% & 29.8s & 2.3$\times$ \\
 & Zero-shot & 697.76 & 0\% & ,  & ,  \\
 & Full fine-tune & 1352.05 & 100\% & 24.0s & 1.9$\times$ \\
 & Top-K & 2021.55 & 48.1\% & 20.9s & 1.6$\times$ \\
\midrule
\multirow{5}{*}{GPT-OSS (20B)}
 & \textbf{Donor} & \textbf{34.56} & 14.6\% & \textbf{2.7m} & ,  \\
 & LoRA & 98.37 & 0.01\% & 76.2m & 28.2$\times$ \\
 & IA$^3$ & 214.99 & 0.0001\% & 76.5m & 28.3$\times$ \\
 & Top-K & 365.83 & 10.6\% & 22.2m & 8.2$\times$ \\
 & Zero-shot & 397.78 & 0\% & ,  & ,  \\
\bottomrule
\end{tabular}
\end{table}

\section{Reproducibility Details}
\label{app:reproducibility}

\subsection{Hardware}

All experiments were conducted on a single compute node equipped with dual NVIDIA GH200 GPUs, each providing 144\,GB of HBM3e memory. The system ran NVIDIA driver version 570.195.03 with CUDA 12.0.

\subsection{Software}

We used the following software versions:

\begin{verbatim}
Python 3.13.9
PyTorch 2.10.0 (CUDA 12.8)
Transformers 4.36.0
PEFT 0.7.1
\end{verbatim}

\subsection{Hyperparameters}

Unless otherwise specified, all experiments used the following default hyperparameters:

\begin{itemize}
 \item Learning rate: $1 \times 10^{-4}$
 \item Batch size: 4-8 (architecture-dependent)
 \item Training epochs: 5
 \item Warmup ratio: 0.1
 \item Maximum sequence length: 512
 \item LoRA rank: 8
 \item LoRA alpha: 16
 \item Random seed: 42
\end{itemize}

\end{document}